%% file: main.tex
\pdfoutput=1
\documentclass[11pt,table]{article}

\usepackage{acl}

\usepackage{times}
\usepackage{latexsym}

\usepackage[T1]{fontenc}
\usepackage{microtype}

\usepackage{amsmath}
\usepackage{graphicx}
\usepackage{amssymb}
\usepackage{booktabs}
\usepackage{multirow}
\usepackage{comment}
\usepackage{siunitx}
\usepackage{enumitem}
\usepackage{arydshln}
\usepackage{color,soul}
\usepackage{tablefootnote}
\usepackage{hyperref}

\makeatletter
\newcommand{\printfnsymbol}[1]{%
  \textsuperscript{\@fnsymbol{#1}}%
}
\makeatother

\usepackage{tabularx,colortbl}

\usepackage{tikz}
\usepackage{collcell}

\usepackage{etoolbox}

\newtoggle{inTableHeader}% Track if still in header of table
\toggletrue{inTableHeader}% Set initial value
\newcommand*{\StartTableHeader}{\global\toggletrue{inTableHeader}}%
\newcommand*{\EndTableHeader}{\global\togglefalse{inTableHeader}}%

\let\OldTabular\tabular%
\let\OldEndTabular\endtabular%
\renewenvironment{tabular}{\StartTableHeader\OldTabular}{\OldEndTabular\StartTableHeader}%

\newcommand*{\MinNumber}{-1.0}%
\newcommand*{\MidNumber}{0.0} %
\newcommand*{\MaxNumber}{1.0}%

\newcommand{\ApplyGradient}[1]{%
  \iftoggle{inTableHeader}{#1}{
    \ifdim #1 pt > \MidNumber pt
        \pgfmathsetmacro{\PercentColor}{max(min(100.0*(#1 - \MidNumber)/(\MaxNumber-\MidNumber),100.0),0.00)} %
        \hspace{-0.33em}\colorbox{yellow!\PercentColor!blue}{#1}
    \else
        \pgfmathsetmacro{\PercentColor}{max(min(100.0*(\MidNumber - #1)/(\MidNumber-\MinNumber),100.0),0.00)} %
        \hspace{-0.33em}\colorbox{blue!\PercentColor!blue}{#1}
    \fi
  }}
\newcolumntype{R}{>{\collectcell\ApplyGradient}c<{\endcollectcell}}

\input{math-com}

\title{SummaReranker: A Multi-Task Mixture-of-Experts \\ Re-ranking Framework for Abstractive Summarization}

\author{Mathieu Ravaut$^\clubsuit$$^\diamondsuit$,
Shafiq Joty$^*$$^\clubsuit$$^\spadesuit$,
Nancy F. Chen$^*$$^\diamondsuit$ \\
$^\clubsuit$ Nanyang Technological University, Singapore\\
$^\diamondsuit$ Institute of Infocomm Research (I$^{2}$R), A$^{*}$STAR, Singapore\\
$^\spadesuit$ Salesforce Research Asia, Singapore\\
\texttt{\{mathieuj001@e.ntu, srjoty@ntu\}.edu.sg}\\
\texttt{nfychen@i2r.a-star.edu.sg}
}

\begin{document}

\maketitle
\def\thefootnote{*}\footnotetext{Equal contribution.}\def\thefootnote{\arabic{footnote}}

\begin{abstract}
Sequence-to-sequence neural networks have recently achieved great success in abstractive summarization, especially through fine-tuning large pre-trained language models on the downstream dataset. These models are typically decoded with beam search to generate a unique summary. However, the search space is very large, and {with the} exposure bias, such decoding is not optimal. In this paper, we show that it is possible to directly train a second-stage model performing \emph{re-ranking} on a set of summary candidates. Our mixture-of-experts SummaReranker learns to select a better candidate and consistently improves the performance of the base model. With a base PEGASUS, we push ROUGE scores by 5.44\% on CNN-DailyMail (47.16 ROUGE-1), 1.31\% on XSum (48.12 ROUGE-1) and 9.34\% on Reddit TIFU (29.83 ROUGE-1), reaching a new state-of-the-art. Our code and checkpoints are available at \url{https://github.com/ntunlp/SummaReranker}.
\end{abstract}

%%%%%%%%%%%%%%%%%%%%%%%%%%%%%%%%%%%%%%%%%%%%%%%%%%%%%%%%%
\section{Introduction}
\label{sec:intro}

\input{Sections/introduction}

%%%%%%%%%%%%%%%%%%%%%%%%%%%%%%%%%%%%%%%%%%%%%%%%%%%%%%%%%
\section{Related Work}
\label{sec:related}

\input{Sections/related}

%%%%%%%%%%%%%%%%%%%%%%%%%%%%%%%%%%%%%%%%%%%%%%%%%%%%%%%%%
\section{Model}
\label{sec:model}

\input{Sections/model}

%%%%%%%%%%%%%%%%%%%%%%%%%%%%%%%%%%%%%%%%%%%%%%%%%%%%%%%%%
\section{Experiments}
\label{sec:exp}

\input{Sections/experiments}

%%%%%%%%%%%%%%%%%%%%%%%%%%%%%%%%%%%%%%%%%%%%%%%%%%%%%%%%%
\section{Discussion}
\label{sec:discussion}

\input{Sections/discussion}

%%%%%%%%%%%%%%%%%%%%%%%%%%%%%%%%%%%%%%%%%%%%%%%%%%%%%%%%%
\section{Conclusion}
\label{sec:conclusion}

\input{Sections/conclusion}

%%%%%%%%%%%%%%%%%%%%%%%%%%%%%%%%%%%%%%%%%%%%%%%%%%%%%%%%%
\section*{Acknowledgements}

\input{Sections/acknowledgements}

%%%%%%%%%%%%%%%%%%%%%%%%%%%%%%%%%%%%%%%%%%%%%%%%%%%%%%%%%
\bibliography{custom}
\bibliographystyle{acl_natbib}

\appendix

\newpage
%%%%%%%%%%%%%%%%%%%%%%%%%%%%%%%%%%%%%%%%%%%%%%%%%%%%%%%%%
\section{Hyper Parameters \& Packages}
\label{sec:appendix_a}

\input{Sections/appendix_a}

\newpage
%%%%%%%%%%%%%%%%%%%%%%%%%%%%%%%%%%%%%%%%%%%%%%%%%%%%%%%%%
\section{Oracle Scores}
\label{sec:appendix_b}

\input{Sections/appendix_b}

\newpage
%%%%%%%%%%%%%%%%%%%%%%%%%%%%%%%%%%%%%%%%%%%%%%%%%%%%%%%%%
\section{Unique Candidates Scores}
\label{sec:appendix_c}

\input{Sections/appendix_c}

\newpage
%%%%%%%%%%%%%%%%%%%%%%%%%%%%%%%%%%%%%%%%%%%%%%%%%%%%%%%%%
\section{Identical Candidates Scores }
\label{sec:appendix_d}

\input{Sections/appendix_d}

\newpage
%%%%%%%%%%%%%%%%%%%%%%%%%%%%%%%%%%%%%%%%%%%%%%%%%%%%%%%%%
\section{Metrics Correlation}
\label{sec:appendix_e}

\input{Sections/appendix_e}

\newpage
%%%%%%%%%%%%%%%%%%%%%%%%%%%%%%%%%%%%%%%%%%%%%%%%%%%%%%%%%
\section{Base Setup Results}
\label{sec:appendix_f}

\input{Sections/appendix_f}

\newpage
%%%%%%%%%%%%%%%%%%%%%%%%%%%%%%%%%%%%%%%%%%%%%%%%%%%%%%%%%
\section{Recall Curves}
\label{sec:appendix_g}

\input{Sections/appendix_g}

% \newpage
%%%%%%%%%%%%%%%%%%%%%%%%%%%%%%%%%%%%%%%%%%%%%%%%%%%%%%%%%
\section{Human Evaluation}
\label{sec:appendix_h}

\input{Sections/appendix_h}

% \newpage
%%%%%%%%%%%%%%%%%%%%%%%%%%%%%%%%%%%%%%%%%%%%%%%%%%%%%%%%%
\section{Candidate Selection}
\label{sec:appendix_i}

\input{Sections/appendix_i}

% \newpage
% %%%%%%%%%%%%%%%%%%%%%%%%%%%%%%%%%%%%%%%%%%%%%%%%%%%%%%%%%
% \section{Abstractiveness}
% \label{sec:appendix_j}

% \input{Sections/appendix_j}

\onecolumn
\newpage
%%%%%%%%%%%%%%%%%%%%%%%%%%%%%%%%%%%%%%%%%%%%%%%%%%%%%%%%%
\section{Speed/Performance Trade-off}
\label{sec:appendix_j}

\input{Sections/appendix_j}

\newpage
%%%%%%%%%%%%%%%%%%%%%%%%%%%%%%%%%%%%%%%%%%%%%%%%%%%%%%%%%
\section{Re-ranking Examples}
\label{sec:appendix_k}

\input{Sections/appendix_k}

\end{document}

%% file: math-com.tex
\usepackage{amsmath,amsfonts,bm}
\usepackage[nameinlink]{cleveref}

\crefformat{section}{\S#2#1#3} % see manual of cleveref, section 8.2.1
\crefname{algorithm}{Alg.}{Algs.}
\crefformat{subsection}{\S#2#1#3}
\Crefname{equation}{Eq.}{Eqs.}
\Crefname{figure}{Fig.}{Figs.}

\makeatletter   
\newcommand{\sveryshortarrow}[1][3pt]{\mathrel{%
    \vcenter{\hbox{\rule[-.5\fontdimen8\scriptfont3]
               {\scriptratio\dimexpr#1\relax}{\fontdimen8\scriptfont3}}}%
   \mkern-4mu\hbox{\let\f@size\sf@size\usefont{U}{lasy}{m}{n}\symbol{41}}}}
\makeatother

\def\eqref#1{equation~\ref{#1}}
\def\1{\bm{1}}

\def\vx{{\bm{x}}}

\def\m1{{\bm{1}}}

\def\mW{{\bm{W}}}

\DeclareMathAlphabet{\mathsfit}{\encodingdefault}{\sfdefault}{m}{sl}
\SetMathAlphabet{\mathsfit}{bold}{\encodingdefault}{\sfdefault}{bx}{n}

\def\gE{{\mathcal{E}}}

\def\gL{{\mathcal{L}}}

\def\gT{{\mathcal{T}}}

\def\sC{{\mathbb{C}}}
\def\sD{{\mathbb{D}}}
\def\sM{{\mathbb{M}}}

\def\sS{{\mathbb{S}}}

\DeclareMathOperator*{\argmax}{arg\,max}

%% file: Sections/introduction.tex
% seq2seq intro
In recent years, sequence-to-sequence neural models have enabled great progress in abstractive summarization \cite{see-etal-2017-get,lin-2021-straight}. In the news domain, they have surpassed the strong LEAD-3 extractive baseline. 
% Initially models like the Pointer-Generator \cite{see-etal-2017-get} were trained from scratch, but 
With the rise of transfer learning since BERT \cite{devlin-etal-2019-bert}, leading approaches typically fine-tune a base pre-trained model that either follows a general text generation training objective like T5 \cite{raffel2019exploring}, BART \cite{lewis-etal-2020-bart}, ERNIE \cite{zhang-etal-2019-ernie} and ProphetNet \cite{qi-etal-2021-prophetnet}, or an objective specifically tailored for summarization like in PEGASUS \cite{zhang2020pegasus}. 

% auto-regressive decoding with beam search is standard, yet not optimal
Most of these sequence-to-sequence models are history-based, where an output sequence is represented as a sequence of decisions and the probability of the sequence is computed as a product of decision probabilities. This is also known as the auto-regressive factorization. To transform the sequence of probabilities into summaries, beam search is commonly used. While auto-regressive decoding with beam search is simple and has many advantages, it can be difficult to encode global constraints such as grammaticality, coherence and factual consistency within this framework, properties that are believed to be useful in discriminating among candidate outputs. If the model starts decoding in a bad direction, mistakes might propagate, carry over the mistake of previous tokens to the generation of new ones, and the model has no way to know that it should adjust the decoding. Furthermore, these models are typically trained with teacher forcing \cite{williams1989learning}, which leads to an inherent discrepancy between training time and inference time known as the exposure bias problem \cite{Bengio-NIPS-15, sun2021alleviating}. 

%, with the number of beams $k$ set to 5, 8 or 10. 

% oracle scores are much better

%%%%%%%%% TABLE 1: oracle scores %%%%%%%%%%

\begin{table}[t!]
\resizebox{\columnwidth}{!}{
\begin{tabular}{lcccccc}
% \textbf{\begin{tabular}[c]{@{}c@{}}Decoding\\ methods\end{tabular}} &
\toprule
\textbf{Decoding methods} &
\textbf{\begin{tabular}[c]{@{}c@{}}\# Summary\\ candidates\end{tabular}} & \textbf{R-1} & \textbf{R-2} & \textbf{R-L} & \textbf{BS} & \textbf{BaS} \\
\midrule
Beam search (top beam) & 1 & 44.23 & 21.48 & 41.21 & 87.39 & -2.78  \\
\midrule
Beam search & 15   & 51.06  & 27.74  & 48.05  & 88.50 & -2.48  \\
Diverse beam search & 15 & \textbf{54.30} & \textbf{30.02} & \textbf{51.33} & \textbf{88.97} & \textbf{-2.40} \\
Top-$k$ sampling & 15  & 52.31 & 27.41 & 49.17 & 88.64 & -2.56 \\
Top-$p$ sampling & 15  & 53.52 & 28.88 & 50.46 & 88.87 & -2.46 \\
\hdashline
{Adding all four methods above} & 60 & \textbf{57.70} & \textbf{33.77} & \textbf{54.72} & \textbf{89.58} & \textbf{-2.25} \\
\bottomrule
\end{tabular}
}
%\vspace{-0.5em}
\caption{ \textbf{Oracle scores} ({maximum scores over all generated candidates}) for four popular decoding methods and five summarization evaluation measures for a base PEGASUS model on \textbf{CNN/DM}. \textbf{R-1/2/L} denotes ROUGE-1/2/L, \textbf{BS} and \textbf{BaS} denote BERTScore and BARTScore, respectively.}
\label{tab:1}
\vspace{-1.0em}
\end{table}

%%%%%%%%%%%%%%%%%%%

Decoding methods such as beam search maintain a list of top-$k$ best candidates, and output a single best one. In the case of beam search, candidates are sorted by decreasing log-probability, and the last $(k-1)$ hypotheses are discarded. However, these $(k-1)$ other hypotheses often contain considerably better sequences in terms of different evaluation measures. This observation holds over other decoding methods: diverse beam search \cite{vijayakumar2016diverse}, top-k sampling \cite{fan-etal-2018-hierarchical} and top-p sampling \cite{holtzman2019curious}. In Table \ref{tab:1}, we illustrate this phenomenon with the \emph{oracle} scores ({maximum scores over the pool of candidates}) for four popular decoding methods and five metrics on the CNN-DailyMail \cite{hermann2015teaching} dataset with a PEGASUS model. The oracle ROUGE-1 scores are up to 10 points higher (+22.8\%) than the top beam baseline. Moreover, oracle gains significantly increase when mixing several generation methods together, reaching an improvement of more than 13 ROUGE-1 points (+30.5\%). Such a gap is larger than the progress made by research in the whole field of neural abstractive summarization in the last five years \cite{nallapati-etal-2016-abstractive,dou-etal-2021-gsum}. This  suggests that current abstractive models are not exploited to their full capacity, calling for better methods to identify the best summary candidate.

%Given an evaluation metric, we refer to the best summary out of $k$ candidates as the \emph{oracle} because finding it means comparing each candidate to the ground truth. 

% motivatinon for SummaReranker
Given this assessment, we investigate whether it is possible to train a \emph{second-stage} summarization model which learns to select the best summary among a set of candidates obtained from a base model and with a decoding process, which itself can potentially involve a set of decoding methods (e.g., beam search variants). This way, the model would recover the gap that separates it with the oracle. This raises the question of what makes a summary candidate the \emph{optimal} one? Admittedly, summarization has been an underconstrained task and its evaluation is complex and remains an active research area    \cite{kryscinski-etal-2019-neural,fabbri2021summeval,koto2021evaluating}. To build a flexible approach, we use a multi-task learning framework based on a mixture-of-experts architecture in order to optimize \emph{jointly} over several measures. 

%The motivation behind this choice is to mimic a real-life scenario where the user specifies which summarization aspect(s) to maximize.

% SummaReranker model
To design a robust re-ranker, we systematically explore the dimensions of summary re-ranking: base model, decoding process, and evaluation measure. 
% 1st advantage: flexibility
Our system, named \emph{SummaReranker}, is flexible and multi-task: it can be trained with any set of evaluation metrics. 
% 2nd advantage: much less expensive than 1st stage models
It is considerably less computationnally expensive to train than the single-stage summarization models that it is plugged on. 
% 3rd advantage: new SOTA 
We apply our system across three different datasets \{CNN-DailyMail, XSum, Reddit TIFU\} and two base models \{PEGASUS, BART\}. Optimizing ROUGE metrics leads to relative performance improvements from 1.31\% to 9.34\% depending on the dataset. It outperforms recently proposed second-stage summarization approaches RefSum \cite{liu-etal-2021-refsum} and SimCLS \cite{liu-liu-2021-simcls} and sets a new state-of-the-art on CNN-DailyMail and XSum \cite{narayan-etal-2018-dont}. 
% Some qualitative results too
We present extensive quantitative results coupled with a qualitative human evaluation. 

%\begin{itemize}
%    \item \textcolor{red}{Most seq2seq models are ``history-based'', in which an output sequence is represented as a sequence of decisions and the probability of the sequence is computed as a product of decision probabilities (also known as autoregressive formulation).}
    
%    \item \textcolor{red}{While this approach is simple and has many advantages, it can be difficult to encode global constraints (e.g., grammaticality, coherence, factual consistency)  within this framework that are believed to be useful in discriminating among candidate outputs.}
    
%    \item \textcolor{red}{However, the $k$-best hypotheses often contain considerably better sequences in terms of different evaluation measures. For example, ... \red{refer to oracle}}
    
%    \item \textcolor{red}{Talk about the dimensions to consider:}
    
%    \begin{itemize}
%        \item \textcolor{red}{how to generate the candidates effectively?}
        
%        \item \textcolor{red}{What measures to use?}
        
%        \item \textcolor{red}{What to optimize?}
        
%        \item \textcolor{red}{What architecture to use for the reranker?}
        
%        \item \textcolor{red}{What information to consider in the reranker (src, candidates)?}
        
%        \item ... 
        
%    \end{itemize}
    
%\end{itemize}

%% file: Sections/related.tex
Re-ranking has been adopted in several branches of NLP for long. 
% parsing
In syntactic parsing, \citet{collins2005discriminative} were the first to employ a re-ranker on the outputs of a base parser, followed by \citet{charniak2005coarse}, who used a Maximum Entropy re-ranker.
% QA
Passage re-ranking is used as the first stage of question-answering systems, to retrieve relevant passages where the answer might lay \cite{kratzwald-feuerriegel-2018-adaptive, nogueira2019passage}. Some recent question-answering models also propose to perform answer re-ranking, to refine the answer selection \cite{kratzwald-etal-2019-rankqa, iyer-etal-2021-reconsider}. 
% MT
Re-ranking has also been used in neural machine translation. Checkpoint reranking \cite{pandramish-sharma-2020-checkpoint} generates several translation candidates with multiple model checkpoints, based on the observation (similar to the one we made in \Cref{sec:intro}) that the oracle across checkpoints is of higher quality than just the last checkpoint. \citet{bhattacharyya-etal-2021-energy} use an energy-based model on top of BERT to select translation candidates with higher BLEU score. 

In abstractive summarization, second-stage approaches such as re-ranking remain underexplored. Recently, RefSum \cite{liu-etal-2021-refsum} defined a second-stage summarization framework which helps address the problem of the train-test distribution mismatch in second-stage models. With a base GSum model \cite{dou-etal-2021-gsum}, the authors reach a 46.18 state-of-the-art ROUGE-1 on CNN-DailyMail. In SimCLS \cite{liu-liu-2021-simcls}, the authors train a second-stage model with contrastive learning, using a ranking loss to select the best summary candidate from a pool of 16 diverse beam search candidates, reaching 46.67 ROUGE-1 on CNN-DailyMail. Our approach differs from RefSum and SimCLS in terms of model architecture and loss function, as well as summary candidate generation process. In contrast with RefSum, we use a single base model, but mix several decoding methods, as our goal is single-model improvement. Unlike SimCLS, we do not use a ranking loss, but directly model the probability that a summary candidate is the best one. To the best of our knowledge, we are the first ones to propose a \emph{multi-task} re-ranking system for abstractive summarization. This enables practitioners to leverage the recent rich literature in automatic abstractive summarization evaluation \cite{lin-2004-rouge, zhang2019bertscore, zhao-etal-2019-moverscore, yuan2021bartscore}.

%% file: Sections/model.tex
%%%%%%%%%%%%%%%%%%%%%%%%
\subsection{Re-ranking Framework}

{Our approach follows the paradigm of second-stage models. Specifically, given a source document $S$, a base model $B$, and a set of decoding methods $\sD$, we get a pool of $m$ summary candidates $\sC = \{C_1, \ldots, C_m\}$. Given an evaluation metric $\mu$ in a set of metrics $\sM$,  we get associated scores for each candidates $\sS_{\mu} = \{ \mu(C_1), \ldots, \mu(C_m) \}$. Our goal is to train a model $f_{\theta}$ parameterized by $\theta$ to explicitly identify the best summary candidate ${C_{\mu}^*}$ according to the metric, which is given by:
\begin{equation} 
{C_{\mu}^*} = \argmax_{C_i \in \sC}~ \{ \mu(C_1), \ldots, \mu(C_m) \}
\end{equation}}
\noindent {We frame this problem as a binary classification. ${C_{\mu}^*}$ is the positive candidate, while other candidates are treated as negative. For a metric $\mu$, the re-ranker $f_{\theta}$ is trained with a binary cross-entropy loss:
\begin{equation}
%\small 
    \gL_{\mu} = - y_i \log p_{\theta}^{\mu}(C_i) - (1-y_{i}) \log (1 - p_{\theta}^{\mu}(C_{i}))
\end{equation}
where $y_i=1$ if $C_i = C_{\mu}^*$, otherwise $y_i=0$}.

{Binary classification has been successfully employed for re-ranking in prior work \cite{nallapati2004discriminative, nogueira2019passage}. While multi-way classification could be an alternative, we noticed that for each generation method, a significant fraction of candidates share the same score for one or several metrics, while it is rare that \emph{all} candidates share the same score (Appendix \ref{sec:appendix_c}-\ref{sec:appendix_d}). Thus, there is not enough signal to distinguish $m$ candidates into $m$ different classes, but enough for two classes.}

To optimize for $N$ different metrics $\sM= \{ \mu_1, \ldots, \mu_N \}$ simultaneously, we use a separate prediction head (tower) for each and we minimize the average over metric losses defined as:
\begin{equation}
%\small
    \gL = \frac{1}{N}\sum\limits_{\mu \in \sM} \gL_{\mu}
\end{equation}

%%%%%%%%%%%%%%%%%%%%%%%%
\subsection{Model Architecture}

We first need to get a good representation of the summary candidate. To use contextual information, we concatenate the source with the candidate, separating the two with a special token: \texttt{\sc{[cls]}} \texttt{Source} \texttt{\sc{[sep]}} \texttt{Candidate}, and feed it to a pre-trained language model. In all experiments, we use RoBERTa-large \cite{liu2019roberta} as encoder. Concatenating the source with the candidate enables RoBERTa to perform cross-attention between the two, which finds parts of the source relevant to the summary candidate. We take the \texttt{\sc{[cls]}} representation from RoBERTa's last layer, and feed it to a multi-layer perceptron (MLP).

%and using a fixed amount of tokens to represent both. 

%%%%%%%%% FIGURE 1: model %%%%%%%%%%

\begin{figure}
    \centering 
    \includegraphics[scale=0.1]{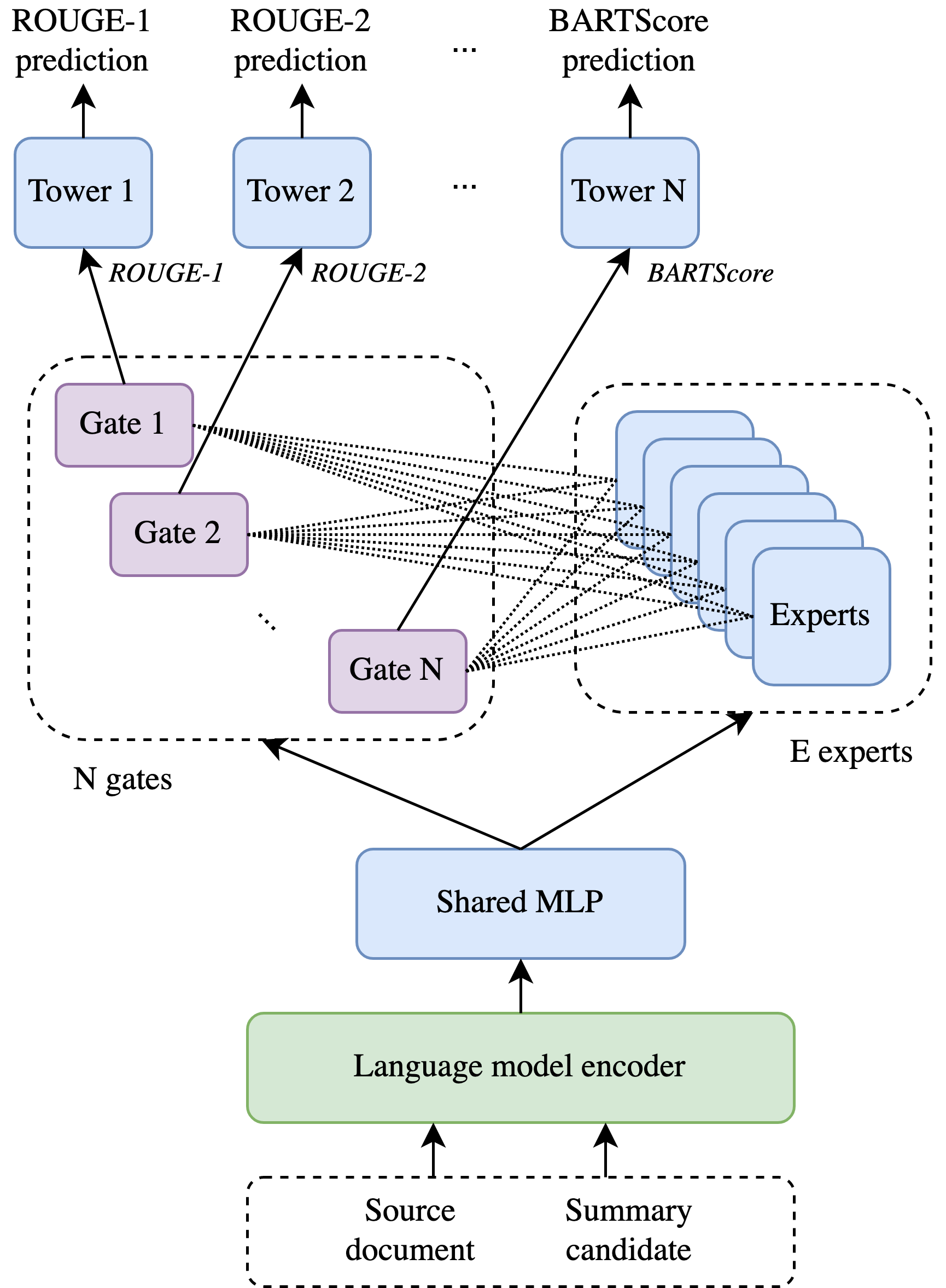}
    \caption{\textbf{SummaReranker model architecture}, optimizing $N$ metrics. The summarization metrics here (\emph{ROUGE-1}, \emph{ROUGE-2}, ...,  \emph{BARTScore}) are displayed as examples.}
    \label{fig:1}
%    \vspace{-1.0em}
\end{figure}

%%%%%%%%%%%%%%%%%%%

{Once we have a joint representation of the source with the candidate (noted $\vx$), we perform multi-task learning in order to optimize for the desired metrics. Since metrics are different, yet may be strongly correlated (e.g., ROUGE variants), we adopt a mixture-of-experts (MoE) architecture. In particular, we follow the sparse MoE approach \cite{shazeer2017outrageously}, which introduces experts dropout. To adapt it to multi-task training, we use the multi-gate approach proposed in \citet{zhao2019recommending}. Given $E$ experts $\gE_{1}, \ldots, \gE_{E}$ and $N$ prediction towers $\gT_{1}, \ldots, \gT_{N}$, the prediction for an input summary representation $\vx$ for a metric $\mu$ indexed by $k \in \{1,  \ldots, N\}$ is:
\begin{equation}
    f_{\theta}^{k}(\vx) = \gT_{k}(\sum\limits_{i=1}^{E} \text{softmax}(\mW_{k}\vx)_{(i)}\gE_{i}(\vx))
\end{equation}
where $\mW_{k}$ is the weight matrix
associated with gate $k$. The corresponding prediction probability is:
\begin{equation}
    p_{\theta}^{\mu} = \text{sigmoid}(f_{\theta}^{k}(\vx))
\end{equation}
Experts are shared across all tasks, and through the softmax gates the model learns how much weight to assign to each expert for each task.}

Our SummaReranker model architecture is shown in \Cref{fig:1}. In practice, the shared bottom MLP consists in two fully-connected layers with ReLU activation \cite{glorot2011deep}. Each expert $\gE_i$ is also a two-layer MLP with ReLU, and each prediction tower $\gT_k$ is a single-layer MLP. We set the number $E$ of experts to be equal to twice the number of tasks ($N$), and the experts dropout to 50\%, so that the effective number of experts being used during training matches $N$. 
{Our model has 370.09 million trainable parameters, representing a slight 4.14\% increase due to the mixture-of-experts compared to the off-the-shelf RoBERTa-large.}

%%%%%%%%%%%%%%%%%%%%%%%%
\subsection{Tackling Training and Inference Gap}

Second-stage learning approaches may suffer from an inherent distribution bias. Indeed, the base model has a different output distribution on the training set than on the validation and test sets. Thus, it is ineffective to train a second-stage model on the training set outputs of the base model. 

%\blue{Needs to be elaborated more for a general reader to understand. Is it better now?} 

To resolve this distribution shift, {we shuffle the training set and randomly split it into equal parts}, then fine-tune a pre-trained model on each half. Then, to build a training set for the re-ranker, we infer with each model on the half that it was not trained on. At testing time, we face two options:
\begin{itemize}[leftmargin=*]
    %\vspace{-0.5em}
    \item \textbf{Base setup}: in this setup, we infer on the test set with \emph{one of the two base models trained on half the training set}, then apply the re-ranker. Since the base models are trained on less data, their performance on the test set worsens. However, we will show that SummaReranker brings improvements which more than compensate this performance drop. 
    %\vspace{-0.6em}
    \item \textbf{Transfer setup}: this setup consists in applying SummaReranker on top of a base model \emph{trained on the whole training set}. Note that SummaReranker is still trained in the same fashion as before. There could be a distribution mismatch in this setting too, since SummaReranker needs to rank summary candidates of a potentially higher quality (generated by a model trained on the full data) than the summaries that it was trained on (generated by a model trained on half the data). Nevertheless, SummaReranker still transfers well and considerably improves the performance of the base model in this {transfer} setup.
\end{itemize}

If $\sD$ is made of multiple decoding methods $\{ \delta_1,...,\delta_{j} \}$, each producing several candidates, the overall candidate set may be large, slowing down inference. Thus, to explore lower-resource inference {setups}, we separate the sets of decoding methods $\sD_{\text{train}}$ and  $\sD_{\text{test}}$ used for training and inference, respectively, and  enforce that $\sD_{\text{test}} \subset \sD_{\text{train}}$.

%\begin{equation}
%\sD_{\text{test}} \subset \sD_{\text{train}}
%\end{equation}

%% file: Sections/experiments.tex
%%%%%%%%%%%%%%%%%%%%%%%%
\subsection{Scope \& Datasets}

Throughout our experiments, we vary all the three dimensions of our re-ranking framework: the base model $B$, the set of decoding methods $\sD$ and the set of scoring metrics $\sM$.

As base models, we use PEGASUS \cite{zhang2020pegasus} and BART \cite{lewis-etal-2020-bart}, each one in their large version, as they are leading summarization models with publicly available checkpoints. We obtain pre-trained and fine-tuned checkpoints from the HuggingFace transformers library \cite{wolf-etal-2020-transformers}. 

For decoding methods ($\sD$), we experiment with beam search (referred to as $1$), diverse beam search (2), top-$k$ sampling (3) and top-$p$ sampling (4). For each decoding method, we set the number of candidates to $15$, as it is close to the maximum which could fit in a standard 11GB RAM GPU when doing generation with PEGASUS-large.

%%%%%%%%% TABLE 2 %%%%%%%%%%

\begin{table}[t]
\renewcommand{\arraystretch}{0}
\setlength{\fboxsep}{3mm} % box size
\setlength{\tabcolsep}{0pt}
\begin{center}
\resizebox{0.86\columnwidth}{!}{
\begin{tabular}{c*{5}{R}}
  & \textbf{R-1} & \textbf{R-2} & \textbf{R-L} & \textbf{BS} & \textbf{BaS} \EndTableHeader\\
  \noalign{\vskip 1mm} 
  \textbf{R-1}~~ & 1.000 & 0.884 & 0.977 & 0.858 & 0.662 \\
  \textbf{R-2}~~ & 0.884 & 1.000 & 0.910 & 0.833 & 0.665 \\
  \textbf{R-L}~~ & 0.977 & 0.910 & 1.000 & 0.855 & 0.669 \\
  \textbf{BS}~~ & 0.858 & 0.833 & 0.855 & 1.000 & 0.682 \\
  \textbf{BaS}~~ & 0.662 & 0.665 & 0.669 & 0.682 & 1.000 \\
\end{tabular}
}
\caption{\textbf{Pearson correlation coefficient} between the five evaluation metrics \{R-1, R-2, R-L, BS, BaS\} for a base PEGASUS with beam search on \textbf{CNN/DM}. \textbf{R-1/2/L} denotes ROUGE-1/2/L, \textbf{BS} and \textbf{BaS} denote BERTScore and BARTScore. 
% {See \Cref{sec:appendix_e} for XSum and Reddit results.}
}
\label{tab:2}
\end{center}
% \vspace{-2.0em}
\end{table}

%%%%%%%%%%%%%%%%%%%

As set of metrics, we first use ROUGE \cite{lin-hovy-2003-automatic}, in its commonly used three flavours of ROUGE-1 (noted \emph{R-1}), ROUGE-2 (noted \emph{R-2}) and ROUGE-L (noted \emph{R-L}) for summarization evaluation. We also leverage recently introduced model based evaluation methods BERTScore (noted \emph{BS}) \cite{zhang2019bertscore} and BARTScore (noted \emph{BaS}) \cite{yuan2021bartscore}, which both rely on contextual word embeddings from pre-trained language models. Thus, our total set of metrics is $\sM$ = \{R-1, R-2, R-L, BS, BaS\}. As seen in \Cref{tab:2}, R-1 and R-L are strongly correlated (Pearson correlation score of 0.977). BARTScore is the least correlated to other metrics, suggesting that it captures aspects complementary to the other four.

We train SummaReranker on the following datasets, covering multiple domains:

\begin{itemize}[leftmargin=*]
    %\vspace{-0.2em}
    \item \textbf{CNN-DailyMail} \cite{hermann2015teaching} contains 93k and 220k articles from the CNN and DailyMail newspapers, respectively. We use the non anonymized version from \cite{see-etal-2017-get}. 
    %\vspace{-0.2em}
    \item \textbf{XSum} \cite{narayan-etal-2018-dont} contains 227k articles from the BBC for years 2010 - 2017. While also in the news domain, XSum is by design significantly more abstractive than CNN/DM and is made of single-sentence summaries. 
     %   \vspace{-0.2em}
    \item \textbf{Reddit TIFU} \cite{kim-etal-2019-abstractive} contains 120k posts from the popular online Reddit forum. As in other summarization works \cite{zhang2020pegasus}, we use the  TIFU-long subset, containing 37k posts. As there is no official split, we build a random 80:10:10 split for training:validation:test. 
\end{itemize}

\noindent We refer to \Cref{tab:3} for statistics on each dataset.

%%%%%%%%% TABLE 3 %%%%%%%%%%

\begin{table}[]
\resizebox{\columnwidth}{!}{
\begin{tabular}{llccccc}
\toprule
\multirow{2}{*}{\textbf{Dataset}} & \multirow{2}{*}{\textbf{Domain}} & \multicolumn{3}{c}{\textbf{\# Data points}} & \multicolumn{2}{c}{\textbf{\# Words}} \\
& & Train & Val & Test & Doc. & Summ. \\
\midrule
CNN/DM & News & 287,113 & 13,368 & 11,490 & 766.56 & 54.78 \\
XSum & News & 204,045 & 11,332 & 11,334 & 414.51 & 22.96 \\
Reddit TIFU & Social media & 33,704 & 4,213 & 4,222 & 385.59 & 20.59 \\
\bottomrule
\end{tabular}
}
% \vspace{-0.5em}
\caption{\textbf{Statistics} of the three datasets.}
% \vspace{-1.0em}
\label{tab:3}
\end{table}

%%%%%%%%%%%%%%%%%%%

%%%%%%%%%%%%%%%%%%%%%%%%
\subsection{Training \& Inference Details}

To help the model better discriminate between candidates, we found that sampling was useful. Specifically, during training, we rank candidates by decreasing sum of normalized scores for the evaluation metrics and keep the top $m_{\text{top}}$ and bottom $m_{\text{bottom}}$ candidates. Thus, training time varies in $\mathcal{O}(m_{\text{top}} + m_{\text{bottom}})$, while inference is in $\mathcal{O}(m)$ as we need to score each candidate. In practice, we found that taking $m_{\text{top}}=1$ and $m_{\text{bottom}}=1$ performed well, on top of decreasing the training time. {This means that at training time, the model only sees two candidates per data point. We scale the pool of candidates that these two are sampled from to \emph{four} decoding methods, totalling 60 summary candidates per source document.}

We train SummaReranker for five epochs. We use the Adafactor optimizer \cite{shazeer2018adafactor}, with maximum learning rate 1e-5, warming up the learning rate linearly over the first 5\% training steps. Training on CNN/DM takes four days on a single RTX 2080 Ti GPU.

{For inference, we need to output a single candidate. After getting predicted probabilities across each metric $\mu \in \sM$, we output the candidate maximizing the sum of predicted probabilities. Note that relaxing inference to allow for a different best candidate for each metric would improve performance, but is not practical. We perform inference with the model checkpoint maximizing the sum of the scores for the metrics on the validation set.}

%%%%%%%%% TABLE 4 %%%%%%%%%%

\begin{table}[t]
\centering 
\resizebox{\columnwidth}{!}{
\begin{tabular}{lcccccc}
\toprule
\textbf{Model} 
& \textbf{\begin{tabular}[c]{@{}c@{}}Model\\ stage\end{tabular}} 
& \textbf{\begin{tabular}[c]{@{}c@{}}Decoding\\ methods ($\sD$)\end{tabular}} 
& \textbf{R-1} 
& \textbf{R-2} 
& \textbf{R-L} 
& \textbf{\begin{tabular}[c]{@{}c@{}}Gain\\ (\%)\end{tabular}} 
\\
\midrule
% \rowcolor{gray!20} 
PEGASUS - 1st half & 1 & \{1\} & 42.23 & 19.62 & 38.90 & \_ \\
PEGASUS - 1st half & 1 & \{2\} & 42.50 & 19.75 & 39.55 & \_\\
% \rowcolor{gray!20} 
PEGASUS - 2nd half & 1 & \{1\} & 42.46 & 19.95 & 39.19 & \_ \\
PEGASUS - 2nd half & 1 & \{2\} & 42.75 & 19.93 & \textbf{39.86} & \_ \\
\hdashline
% \rowcolor{gray!20} 
BART - 1st half & 1 & \{1\} & 42.79 & 20.25 & 39.66 & \_ \\
BART - 1st half & 1 & \{2\} & 40.70 & 18.99 & 37.88 & \_ \\
% \rowcolor{gray!20} 
BART - 2nd half & 1 & \{1\} & \textbf{42.93} & \textbf{20.36} & 39.73 & \_ \\
BART - 2nd half & 1 & \{2\} & 41.93 & 19.79 & 39.06 & \_ \\
\midrule
% \rowcolor{gray!20} 
PEGASUS - 1st half \textbf{+ SR} & 2 & \{1\} & 44.02 & 20.97 & 40.68 & 5.23 \\
PEGASUS - 1st half \textbf{+ SR} & 2 & \{2\} & \textbf{45.66} & 21.31 & 42.51 & 7.61 \\
% \rowcolor{gray!20} 
PEGASUS - 2nd half \textbf{+ SR} & 2 & \{1\} & 44.11 & 21.08 & 40.82 & 4.57 \\
PEGASUS - 2nd half \textbf{+ SR} & 2 & \{2\} & 45.73 & 21.31 & \textbf{42.62} & 6.94 \\
\hdashline
% \rowcolor{gray!20} 
BART - 1st half \textbf{+ SR} & 2 & \{1\} & 44.23 & 21.23 & 41.09 & 3.94 \\
BART - 1st half \textbf{+ SR} & 2 & \{2\} & 45.05 & 21.47 & 42.12 & 11.65 \\
% \rowcolor{gray!20} 
BART - 2nd half \textbf{+ SR} & 2 & \{1\} & 44.51 & 21.52 & 41.29 & 4.44 \\
BART - 2nd half \textbf{+ SR} & 2 & \{2\} & 45.61 & \textbf{21.78} & \textbf{42.62} & 9.32 \\
\midrule
PEGASUS - 1st half \textbf{+ SR} & 2 & \{1, 2\} & 46.12 & 21.97 & 42.84 & 9.36 \\
PEGASUS - 2nd half \textbf{+ SR} & 2 & \{1, 2\} & \textbf{46.19} & 22.02 & \textbf{42.92} & 8.70 \\
\hdashline
BART - 1st half \textbf{+ SR} & 2 & \{1, 2\} & 45.76 & 22.14 & 42.71 & 7.99 \\
BART - 2nd half \textbf{+ SR} & 2 & \{1, 2\} & 45.96 & \textbf{22.18} & 42.88 & 7.98 \\
\bottomrule
\end{tabular}
}
\caption{\textbf{Base setup results} for SummaReranker applied to PEGASUS and BART on the \textbf{CNN/DM} dataset. \textbf{SR} refers to SummaReranker. {\textbf{Decoding method \{1\}} is beam search, \textbf{\{2\}} is diverse beam search}. Best scores for each type of model are in bold. \textbf{Gain} represents the mean relative gain over \{R-1, R-2, R-L\} {compared to} the best decoding method. 
% {Refer to \Cref{sec:appendix_f} for the results on XSum and Reddit TIFU.}
}
\label{tab:4}
% \vspace{-1.5em}
\end{table}

%%%%%%%%%%%%%%%%%%%

%%%%%%%%% TABLE 5 %%%%%%%%%%

\begin{table*}
\centering 
\resizebox{\textwidth}{!}{
\begin{tabular}{lccccc|cccccc}
\toprule 
& & \multicolumn{2}{c}{\textbf{Decoding methods}} & & & \multicolumn{5}{c}{\textbf{Evaluation metrics}} \\
\textbf{Model} 
& \textbf{\begin{tabular}[c]{@{}c@{}}Model\\ stage\end{tabular}} 
& $\sD_{\text{train}}$
& $\sD_{\text{test}}$
% & \textbf{\begin{tabular}[c]{@{}c@{}} \# Summary\\ candidates \end{tabular}} 
& $m$ 
& \textbf{\begin{tabular}[c]{@{}c@{}}Optimized\\  Metrics ($\sM$)\end{tabular}} 
& \textbf{R-1} 
& \textbf{R-2} 
& \textbf{R-L} 
& \textbf{BS} 
& \textbf{BaS}   
& \textbf{\begin{tabular}[c]{@{}c@{}}Gain\\  (\%)\end{tabular}} 
\\
\midrule
% \rowcolor{gray!20} 
PEGASUS \cite{zhang2020pegasus} & 1 & \{1\} & \{1\} & 8 & \_ & 44.16 & 21.56 & 41.30 & \_ & \_ & \_ \\
% \rowcolor{gray!20} 
PEGASUS - \emph{our setup} & 1 & \{1\} & \{1\} & 15 & \_ & 44.23 & 21.48 & 41.21 & 87.39 & -2.78 & \_ \\
PEGASUS - \emph{our setup} & 1 & \{2\} & \{2\} & 15 & \_ & 44.56 & 20.90 & 41.58 & 87.36 & -2.81 & \_ \\
% \rowcolor{gray!20} 
BART \cite{lewis-etal-2020-bart} & 1 & \{1\} & \{1\} & 5 & \_ & 44.16 & 21.28 & 40.90 & \_ & \_ & \_ \\
% \rowcolor{gray!20} 
BART - \emph{our setup} & 1 & \{1\} & \{1\} & 15 & \_ & 43.28 & 20.44 & 40.06 & 87.78 & -2.48 & \_ \\
BART - \emph{our setup} & 1 & \{2\} & \{2\} & 15 & \_ & 44.48 & 21.21 & 41.60 & \textbf{88.11} & \textbf{-2.33} & \_ \\
\hdashline
% \rowcolor{gray!20} 
BART + R3F \cite{aghajanyan2020better} & 1 & \{1\} & \{1\} & 5 & \_ & 44.38 & 21.53 & 41.17 & \_ & \_ & \_ \\
% \rowcolor{gray!20} 
GSum \cite{dou-etal-2021-gsum} & 1 & \{1\} & \{1\} & 4 & \_ & \textbf{45.94} & \textbf{22.32} & \textbf{42.48} & \_ & \_ & \_ \\
\midrule
GSum + RefSum \cite{liu-etal-2021-refsum} & 2 & \{1\} & \{1\} & 4 & \_ & 46.18 & 22.36 & 42.91 & \_ & \_ & \_ \\
BART + SimCLS \cite{liu-liu-2021-simcls} & 2 & \{2\} & \{2\} & 16 & \_ & 46.67 & 22.15 & 43.54 & 66.14 & \_ & \_ \\
\hdashline
% PEGASUS + SR & 2 & \{2\} & \{2\} & 15 & \{mean R\} & 46.86$^{\dagger}$ & 22.04$^{\dagger}$ & 43.59$^{\dagger}$ & 87.67$^{\dagger}$ & -2.73$^{\dagger}$ & 4.82 \\
% \rowcolor{gray!20} 
PEGASUS \textbf{+ SR} & 2 & \{1\} & \{1\} & 15 & \{R-1, R-2, R-L\} & \cellcolor{gray!25} 45.56$^{\dagger}$ & \cellcolor{gray!25} 22.23$^{\dagger}$ & \cellcolor{gray!25} 42.46$^{\dagger}$ & 87.60$^{\dagger}$ & -2.74$^{\dagger}$ & 3.18 \\
PEGASUS \textbf{+ SR} & 2 & \{2\} & \{2\} & 15 & \{R-1, R-2, R-L\} & \cellcolor{gray!25} \textbf{46.86}$^{\dagger}$ & \cellcolor{gray!25} 22.01$^{\dagger}$ & \cellcolor{gray!25} \textbf{43.59}$^{\dagger}$ & 87.66$^{\dagger}$ & -2.73$^{\dagger}$ & 5.10 \\
% \rowcolor{gray!20} 
PEGASUS \textbf{+ SR} & 2 & \{1, 2\} & \{1\} & 15 & \{R-1, R-2, R-L\} & \cellcolor{gray!25} 46.13$^{\dagger}$ & \cellcolor{gray!25} \textbf{22.61}$^{\dagger}$ & \cellcolor{gray!25} 42.94$^{\dagger}$ & 87.67$^{\dagger}$ & -2.72$^{\dagger}$ & 4.59 \\
PEGASUS \textbf{+ SR} & 2 & \{1, 2\} & \{2\} & 15 & \{R-1, R-2, R-L\} & \cellcolor{gray!25} 46.83$^{\dagger}$ & \cellcolor{gray!25} 21.88$^{\dagger}$ & \cellcolor{gray!25} 43.55$^{\dagger}$ & 87.63$^{\dagger}$ & -2.74$^{\dagger}$ & 4.84 \\
% \rowcolor{gray!20} 
BART \textbf{+ SR} & 2 & \{1\} & \{1\} & 15 & \{R-1, R-2, R-L\} & \cellcolor{gray!25} 44.60$^{\dagger}$ & \cellcolor{gray!25} 21.38$^{\dagger}$ & \cellcolor{gray!25} 41.36$^{\dagger}$ & 88.03$^{\dagger}$ & -2.40$^{\dagger}$ & 3.63 \\
BART \textbf{+ SR} & 2 & \{2\} & \{2\} & 15 & \{R-1, R-2, R-L\} & \cellcolor{gray!25} 46.47$^{\dagger}$ & \cellcolor{gray!25} 22.17$^{\dagger}$ & \cellcolor{gray!25} 43.45$^{\dagger}$ & 88.43$^{\dagger}$ & -2.19$^{\dagger}$ & 4.48 \\
% \rowcolor{gray!20} 
BART \textbf{+ SR} & 2 & \{1, 2\} & \{1\} & 15 & \{R-1, R-2, R-L\} & \cellcolor{gray!25} 45.08$^{\dagger}$ & \cellcolor{gray!25} 21.79$^{\dagger}$ & \cellcolor{gray!25} 41.85$^{\dagger}$ & 88.13$^{\dagger}$ & -2.37$^{\dagger}$ & 5.08 \\
BART \textbf{+ SR} & 2 & \{1, 2\} & \{2\} & 15 & \{R-1, R-2, R-L\} & \cellcolor{gray!25} 46.50$^{\dagger}$ & \cellcolor{gray!25} 22.15$^{\dagger}$ & \cellcolor{gray!25} 43.50$^{\dagger}$ & \textbf{88.45}$^{\dagger}$ & \textbf{-2.18}$^{\dagger}$ & 4.51 \\
\hdashline
% PEGASUS + SR & 2 & \{1, 2\} &  \{1, 2\} & 30 & \{mean R\} & 47.03$^{\dagger}$ & 22.47$^{\dagger}$ & 43.75$^{\dagger}$ & 87.73$^{\dagger}$ & -2.71$^{\dagger}$ & 5.80 \\
PEGASUS \textbf{+ SR} (\textbf{new SOTA}) & 2 & \{1, 2\} & \{1, 2\} & 30 & \{R-1, R-2, R-L\} & \cellcolor{gray!25} \textbf{47.16}$^{\dagger}$ & \cellcolor{gray!25} \textbf{22.55}$^{\dagger}$ & \cellcolor{gray!25} \textbf{43.87}$^{\dagger}$ & 87.74$^{\dagger}$ & -2.71$^{\dagger}$ & \textbf{5.44} \\
PEGASUS \textbf{+ SR} & 2 & \{1, 2\} & \{1, 2\} & 30 & \{BS, BaS\} & 45.00$^{\dagger}$ & 20.90 & 41.93$^{\dagger}$ & \cellcolor{gray!25} 87.56$^{\dagger}$ & \cellcolor{gray!25} -2.55$^{\dagger}$ & 4.23 \\
PEGASUS \textbf{+ SR} & 2 & \{1, 2\} & \{1, 2\} & 30 & \{R-1, R-2, R-L, BS, BaS\} & \cellcolor{gray!25} 46.59$^{\dagger}$ & \cellcolor{gray!25} 22.41$^{\dagger}$ & \cellcolor{gray!25} 43.45$^{\dagger}$ & \cellcolor{gray!25} 87.77$^{\dagger}$ & \cellcolor{gray!25} -2.58$^{\dagger}$ & 4.39 \\
BART \textbf{+ SR} & 2 & \{1, 2\} & \{1, 2\} & 30 & \{R-1, R-2, R-L\} & \cellcolor{gray!25} 46.62$^{\dagger}$ & \cellcolor{gray!25} 22.39$^{\dagger}$ & \cellcolor{gray!25} 43.59$^{\dagger}$ & \textbf{88.47}$^{\dagger}$ & -2.18$^{\dagger}$ & 5.05 \\
BART \textbf{+ SR} & 2 & \{1, 2\} & \{1, 2\} & 30 & \{BS, BaS\} & 44.90$^{\dagger}$ & 20.85 & 42.03$^{\dagger}$ & \cellcolor{gray!25} 88.28$^{\dagger}$ & \cellcolor{gray!25} \textbf{-2.05}$^{\dagger}$ & 6.11 \\
BART \textbf{+ SR} & 2 & \{1, 2\} & \{1, 2\} & 30 & \{R-1, R-2, R-L, BS, BaS\} & \cellcolor{gray!25} 45.96$^{\dagger}$ & \cellcolor{gray!25} 21.79$^{\dagger}$ & \cellcolor{gray!25} 43.01$^{\dagger}$ & \cellcolor{gray!25} 88.44$^{\dagger}$ & \cellcolor{gray!25} -2.09$^{\dagger}$ & 4.03  \\
\hdashline
PEGASUS \textbf{+ SR} & 2 & \{1, 2, 3, 4\} &  \{1, 2, 3, 4\} & 60 & \{R-1, R-2, R-L\} & \cellcolor{gray!25} \textbf{47.04}$^{\dagger}$ & \cellcolor{gray!25} \textbf{22.32}$^{\dagger}$ & \cellcolor{gray!25} \textbf{43.72}$^{\dagger}$ & \textbf{87.69}$^{\dagger}$ & \textbf{-2.74}$^{\dagger}$ & \_ \\
\bottomrule
\end{tabular}
}
% \vspace{-0.5em}
\caption{\textbf{Transfer setup results on CNN/DM}. \textbf{SR} refers to SummaReranker, $m$ refers to the number of summary candidates, \textbf{BS} and \textbf{BaS} to BERTScore and BARTScore, respectively. Best scores for each type of model (single stage, second-stage) are in bold. $^{\dagger}$ marks are results significantly better than the base model counterpart among metrics that SummaReranker was optimized for. Results for optimized metrics are shaded. \textbf{Gain} represents the mean relative gain over optimized metrics.}
\label{tab:5}
% \vspace{-1.0em}
\end{table*}

%%%%%%%%%%%%%%%%%%%

% \vspace{-0.5em}
%%%%%%%%%%%%%%%%%%%%%%%%
\subsection{Base Setup Results}

First, we investigate how our model performs in the base setup described in  \Cref{sec:model}. We apply SummaReranker on top of PEGASUS and BART models fine-tuned on each half. For each model, we decode using beam search (1) and diverse beam search (2). The latter performs better for PEGASUS, while the former is better for BART. We then apply SummaReranker optimized jointly for R-1, R-2, and R-L on top of each of the two base models, for each decoding method, and finally when using both decoding methods. Results are shown in \Cref{tab:4}. 

SummaReranker improves a base PEGASUS by 4.57\% to 7.21\% with 15 candidates, and 8.70\% to 9.36\% with 30 candidates. With BART, SummaReranker improves by 3.94\% to 11.65\% with 15 candidates, and 7.98\% with 30 candidates. When using several decoding methods, we compare the re-ranker performance with the best baseline among decoding methods. Notably, with SummaReranker, PEGASUS and BART models trained on 50\% of the training set now surpass their counterparts trained on \emph{the whole} training set, achieving 46.19 R-1 with PEGASUS and 45.96 R-1 with BART. This is better than GSum \cite{dou-etal-2021-gsum}, the best {reported} summarization model on CNN/DM.

%%%%%%%%% TABLE 6 %%%%%%%%%%

\begin{table*}[t]
\resizebox{0.99\textwidth}{!}{
\begin{tabular}{lcccc|cccccc|cccccc}
\toprule
& & \multicolumn{2}{c}{\textbf{Decoding methods}} & & \multicolumn{6}{c}{\textbf{XSum}} & \multicolumn{6}{c}{\textbf{Reddit TIFU}} \\
\textbf{Model}    
& \textbf{\begin{tabular}[c]{@{}c@{}}Model\\ stage\end{tabular}}
& \textbf{$\sD_{\text{train}}$} 
& \textbf{$\sD_{\text{test}}$} 
% & \textbf{\begin{tabular}[c]{@{}c@{}}\# Summary\\ candidates\end{tabular}} 
& $m$
& \textbf{R-1} 
& \textbf{R-2}
& \textbf{R-L} 
& \textbf{BS}
& \textbf{BaS} 
& \textbf{\begin{tabular}[c]{@{}c@{}}Gain\\  (\%)\end{tabular}} 
& \textbf{R-1} 
& \textbf{R-2} 
& \textbf{R-L} 
& \textbf{BS} 
& \textbf{BaS} 
& \textbf{\begin{tabular}[c]{@{}c@{}}Gain\\  (\%)\end{tabular}} 
\\
\midrule
% \rowcolor{gray!20} 
PEGASUS \cite{zhang2020pegasus} & 1 & \{1\} & \{1\} & 8 & 47.21 & 24.56 & 39.25 & \_ & \_ & \_ & \emph{26.63} & \emph{9.01} & \emph{21.60} & \_ & \_ & \_ \\
% \rowcolor{gray!20} 
PEGASUS - \emph{our setup} & 1 & \{1\} & \{1\} & 15 & \textbf{47.33} & \textbf{24.75} & \textbf{39.43} & \textbf{92.01} & \textbf{-1.92} & \_ & 26.28 & 9.01 & 21.52 & 87.34 & \textbf{-3.46} & \_ \\
PEGASUS - \emph{our setup} & 1 & \{2\} & \{2\} & 15 & 46.78 & 23.77 & 38.70 & 91.94 & -2.00 & \_ & 25.67 & 8.07 & 20.97 & 87.47 & -3.48 & \_ \\
% \rowcolor{gray!20} 
BART \cite{lewis-etal-2020-bart} & 1 & \{1\} & \{1\} & 5 & 45.14 & 22.27 & 37.25 & \_ & \_ & \_ & \_ & \_ & \_ & \_ & \_ & \_ \\
% \rowcolor{gray!20} 
BART - \emph{our setup} & 1 & \{1\} & \{1\} & 15 & 45.24 & 22.28 & 37.21 & 91.58 & -1.97 & \_ & \textbf{27.42} & \textbf{9.53} & \textbf{22.10} & 87.43 & -3.78 & \_ \\
BART - \emph{our setup} & 1 & \{2\} & \{2\} & 15 & 44.15 & 20.84 & 35.88 & 91.51 & -2.08 & \_ & 25.43 & 8.27 & 20.79 & \textbf{87.48} & -4.19 & \_ \\
\hdashline
% \rowcolor{gray!20} 
BART + R3F \cite{aghajanyan2020better} & 1 & \{1\} & \{1\} & 5 & \_ & \_ & \_ & \_ & \_ & \_ & \emph{30.31} & \emph{10.98} & \emph{24.74} & \_ & \_ & \_ \\
% \rowcolor{gray!20} 
\midrule
GSum + RefSum \cite{liu-etal-2021-refsum} & 2 & \{1\} & \{1\} & 4 & 47.45 & 24.55 & 39.41 & \_ & \_ & \_ & \_ & \_ & \_ & \_ & \_ & \_ \\
PEGASUS + SimCLS \cite{liu-liu-2021-simcls} & 2 & \{2\} & \{2\} & 16 & 47.61 & 24.57 & 39.44 & 69.81 & \_ & \_ & \_ & \_ & \_ & \_ & \_ & \_ \\
\hdashline
% \rowcolor{gray!20} 
PEGASUS \textbf{+ SR} (\textbf{new XSum SOTA}) & 2 & \{1, 2\} & \{1\} & 15 & \cellcolor{gray!25} \textbf{48.12}$^{\dagger}$ & \cellcolor{gray!25} \textbf{24.95} & \cellcolor{gray!25} \textbf{40.00}$^{\dagger}$ & \textbf{92.14}$^{\dagger}$ & \textbf{-1.90}$^{\dagger}$ & \textbf{1.31} & \cellcolor{gray!25} \textbf{29.57}$^{\dagger}$ & \cellcolor{gray!25} 9.70$^{\dagger}$ & \cellcolor{gray!25} \textbf{23.29}$^{\dagger}$ & 87.63$^{\dagger}$ & \textbf{-3.34}$^{\dagger}$ & 9.47 \\
PEGASUS \textbf{+ SR} & 2 & \{1, 2\} & \{2\} & 15 & \cellcolor{gray!25} 47.04 & \cellcolor{gray!25} 23.27 & \cellcolor{gray!25} 38.55 & 91.98 & -2.01 & -0.65 & \cellcolor{gray!25} 28.71$^{\dagger}$ & \cellcolor{gray!25} 8.73$^{\dagger}$ & \cellcolor{gray!25} 22.79$^{\dagger}$ & \textbf{87.84}$^{\dagger}$ & -3.42$^{\dagger}$ & 9.57 \\
% \rowcolor{gray!20} 
BART \textbf{+ SR} & 2 & \{1, 2\} & \{1\} & 15 & \cellcolor{gray!25} 45.79$^{\dagger}$ & \cellcolor{gray!25} 22.17 & \cellcolor{gray!25} 37.31 & 91.69$^{\dagger}$ & -1.97 & 0.33 & \cellcolor{gray!25} 28.99$^{\dagger}$ & \cellcolor{gray!25} \textbf{9.82} & \cellcolor{gray!25} 22.96$^{\dagger}$ & 87.53 & -3.78 & 4.22 \\
BART \textbf{+ SR} & 2 & \{1, 2\} & \{2\} & 15 & \cellcolor{gray!25} 44.39 & \cellcolor{gray!25} 20.35 & \cellcolor{gray!25} 35.66 & 91.51 & -2.16 & -0.81 & \cellcolor{gray!25} 28.04$^{\dagger}$ & \cellcolor{gray!25} 8.66 & \cellcolor{gray!25} 22.41$^{\dagger}$ & 87.73$^{\dagger}$ & -3.91$^{\dagger}$ & 7.59 \\
\hdashline
PEGASUS \textbf{+ SR} (\textbf{best Reddit TIFU score}) & 2 & \{1, 2\} & \{1, 2\} & 30 & \cellcolor{gray!25} \textbf{47.72} & \cellcolor{gray!25} \textbf{24.16} & \cellcolor{gray!25} \textbf{39.42} & \textbf{92.10}$^{\dagger}$ & \textbf{-1.94} & -0.53 & \cellcolor{gray!25} \textbf{29.83}$^{\dagger}$ & \cellcolor{gray!25} \textbf{9.50}$^{\dagger}$ & \cellcolor{gray!25} \textbf{23.47}$^{\dagger}$ & \textbf{87.81}$^{\dagger}$ & \textbf{-3.33}$^{\dagger}$ & \textbf{9.34} \\ 
BART \textbf{+ SR} & 2 & \{1, 2\} & \{1, 2\} & 30 & \cellcolor{gray!25} 45.32 & \cellcolor{gray!25} 21.46 & \cellcolor{gray!25} 36.64 & 91.64 & -2.04 & -1.68 & \cellcolor{gray!25} 28.92$^{\dagger}$ & \cellcolor{gray!25} 9.16 & \cellcolor{gray!25} 22.87$^{\dagger}$ & 87.70$^{\dagger}$ & -3.83$^{\dagger}$ & 1.69 \\ 
\bottomrule
\end{tabular}
}
% \vspace{-0.5em}
\caption{\textbf{Transfer setup results on XSum and Reddit TIFU}. \textbf{SR} refers to SummaReranker, $m$ refers to the number of summary candidates, \textbf{BS} and \textbf{BaS} to BERTScore and BARTScore, respectively. Best scores for each type of model (single stage, second-stage) are in bold. $^{\dagger}$ marks are results significantly better than the base model counterpart among metrics that SummaReranker was optimized for. Results for optimized metrics are shaded. \textbf{Gain} represents the mean relative gain over optimized metrics. Reddit TIFU results in italic are not directly comparable due to a different data split.}
\label{tab:6}
% \vspace{-1.5em}
\end{table*}

%%%%%%%%%%%%%%%%%%%

%%%%%%%%%%%%%%%%%%%%%%%%
\subsection{Transfer Setup Results}

Next, we look at how SummaReranker performs in the transfer setup. That means, we apply it on top of PEGASUS and BART models fine-tuned on the entire dataset, using public checkpoints. We also include R3F \cite{aghajanyan2020better} and GSum \cite{dou-etal-2021-gsum} in our single-stage model comparison. In terms of second-stage approaches, we compare SummaReranker with RefSum \cite{liu-etal-2021-refsum} and SimCLS \cite{liu-liu-2021-simcls}. Note that SummaReranker is trained as usual, on the outputs of two base models each trained on 50\%.

% Results on the CNN/DM dataset are shown in \Cref{tab:5}. 
We first optimize for ROUGE metric \{R-1, R-2, R-L\} with multi-task training on CNN/DM (\Cref{tab:5}). 
% With a single decoding method and 15 summary candidates, SummaReranker places PEGASUS and BART on par with SimCLS. 
With two decoding methods, PEGASUS + SummaReranker sets a new state of the art on CNN/DM with 47.16 R-1, 22.55 R-2 and 43.87 R-L, corresponding to gains of 2.60/1.65/2.29 R-1/2/L or +5.44\% from our diverse beam search baseline. As expected, the relative gains in transfer setup are lower than in base setup. Next, we optimize model-based metrics, and note the difficulty in improving BERTScore, compared to BARTScore. Optimizing jointly ROUGE and model-based metrics improves all metrics, but does not match the results when training only ROUGE. Interestingly, performance gains saturate when adding two extra decoding methods (top-$k$ and top-$p$ sampling), despite gains in the oracle scores observed in \Cref{tab:1}. 
% We note that performance could be further improved with a GSum base model.

To assert statistical significance of performance gains, we perform a t-test between SummaReranker scores and scores from the base model with \emph{each} of the decoding methods being used, and mark with $\dagger$ results where the $p$-value is smaller than 0.05 for \emph{all} these decoding methods. 

\begin{figure}
    \centering 
    \includegraphics[width=0.99\columnwidth]{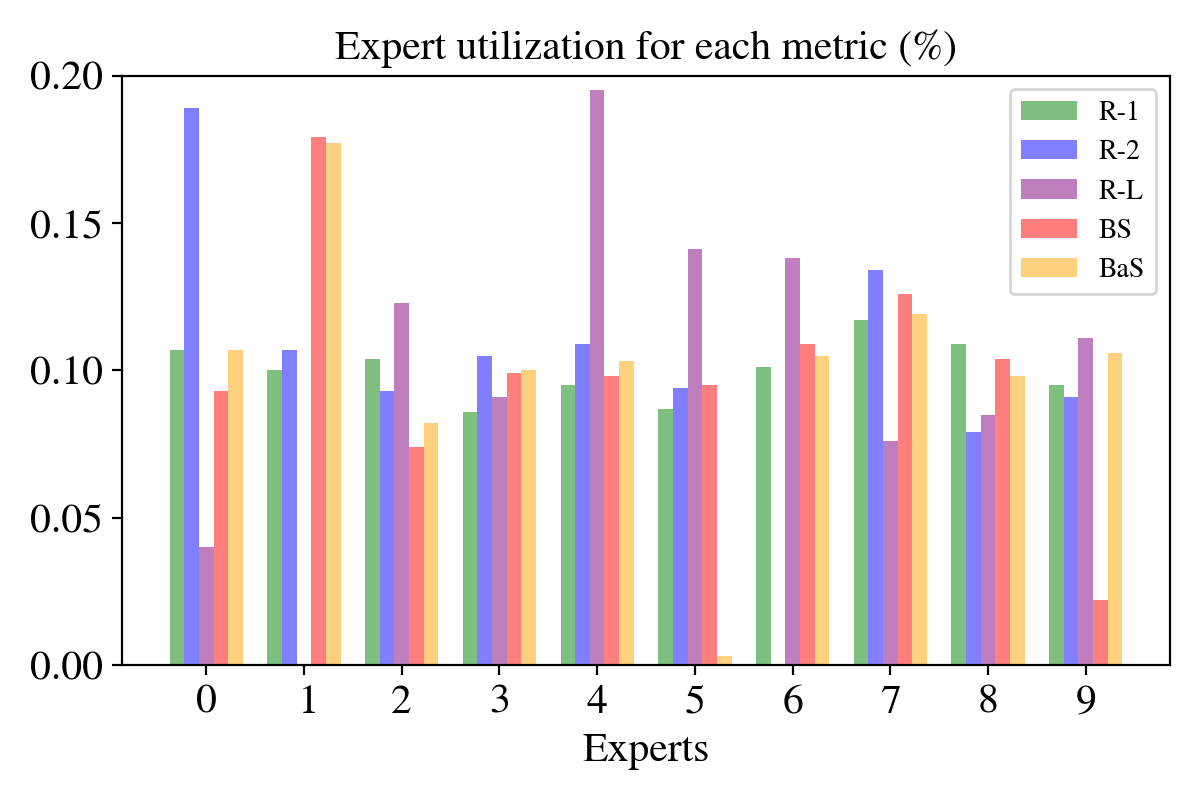}
    % \vspace{-2.0em}
    \caption{\textbf{Expert utilization} for a base PEGASUS with SummaReranker optimized with \{R-1, R-2, R-L, BS, BaS\} on \textbf{CNN/DM}, with 10 experts.}
    \label{fig:2}
    % \vspace{-1.0em}
\end{figure}

We also show experts utilization (obtained with softmax weights from the gates) for the model optimized on all five metrics in \Cref{fig:2}. Notably, some experts specialize in certain metrics (for instance, expert 0 on R-2 and expert 4 on R-L). 

Then, we apply SummaReranker on XSum and Reddit TIFU, as shown in \Cref{tab:6}. We train SummaReranker using the three ROUGE metrics \{R-1, R-2, R-L\} as objective, and decoding methods \{beam search, diverse beam search\} to generate the candidates. On XSum, SummaReranker improves a base PEGASUS with beam search candidates by 1.31\%, setting a new state-of-the-art of 48.12/24.95/40.00 R-1/2/L. On Reddit TIFU, we improve a base PEGASUS with beam search and diverse beam search (30 candidates) by 9.34\%, reaching 29.83/9.50/23.47 R-1/2/L, and a base BART with beam search by 4.22\%, reaching 28.99/9.82/22.96 R-1/2/L. Across datasets, training on a combination of beam search and diverse beam search candidates is consistently effective.

%%%%%%%%% FIGURE 2 %%%%%%%%%%

\vspace{-1.0em}

%%%%%%%%% FIGURE 3 %%%%%%%%%%

\begin{figure}
    \centering 
    \includegraphics[width=0.99\columnwidth]{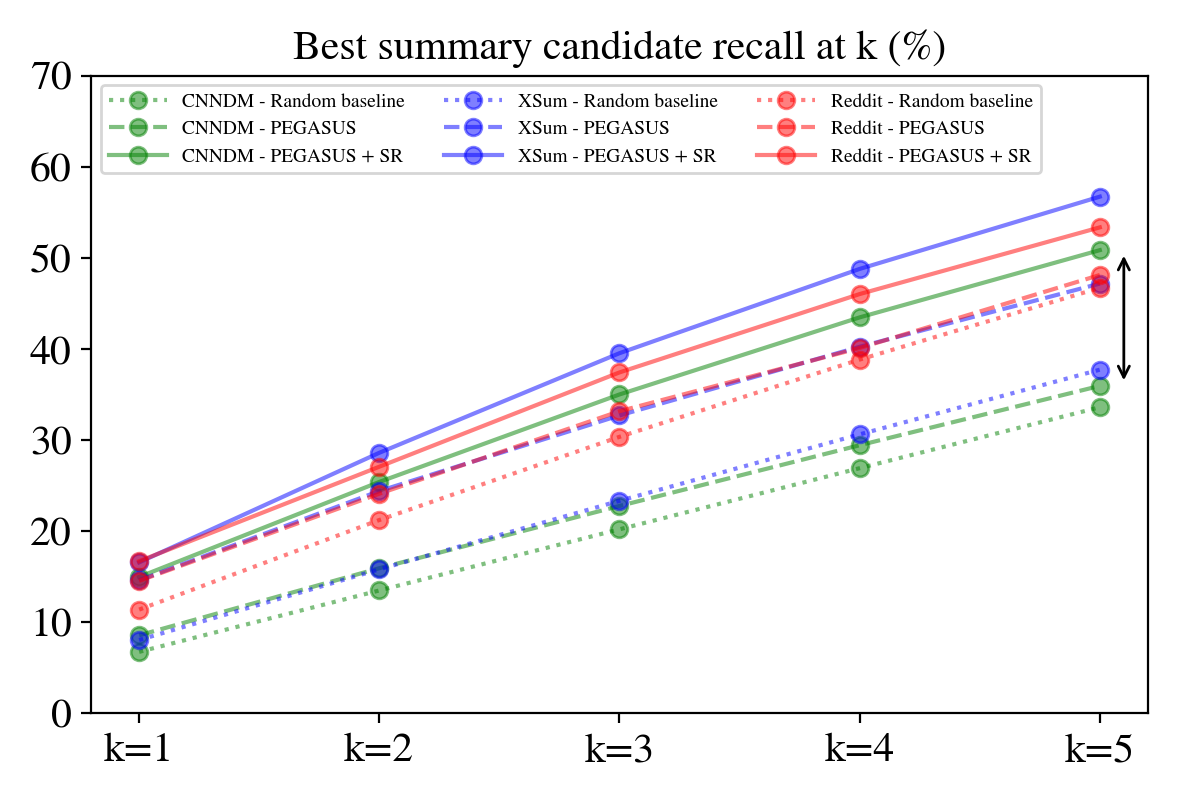}
    % \vspace{-2em}
    \caption{\textbf{Best summary candidate recall} with 15 diverse beam search candidates for PEGASUS on all three datasets. SR denotes SummaReranker. Dotted lines are random baselines, and dashed lines correspond to the base PEGASUS.}
    \label{fig:3}
    % \vspace{-2.5em}
\end{figure}

%%%%%%%%%%%%%%%%%%%

%%%%%%%%% FIGURE 4: example %%%%%%%%%%

\begin{figure*}[t]
    \centering 
    \includegraphics[width=0.99\textwidth]{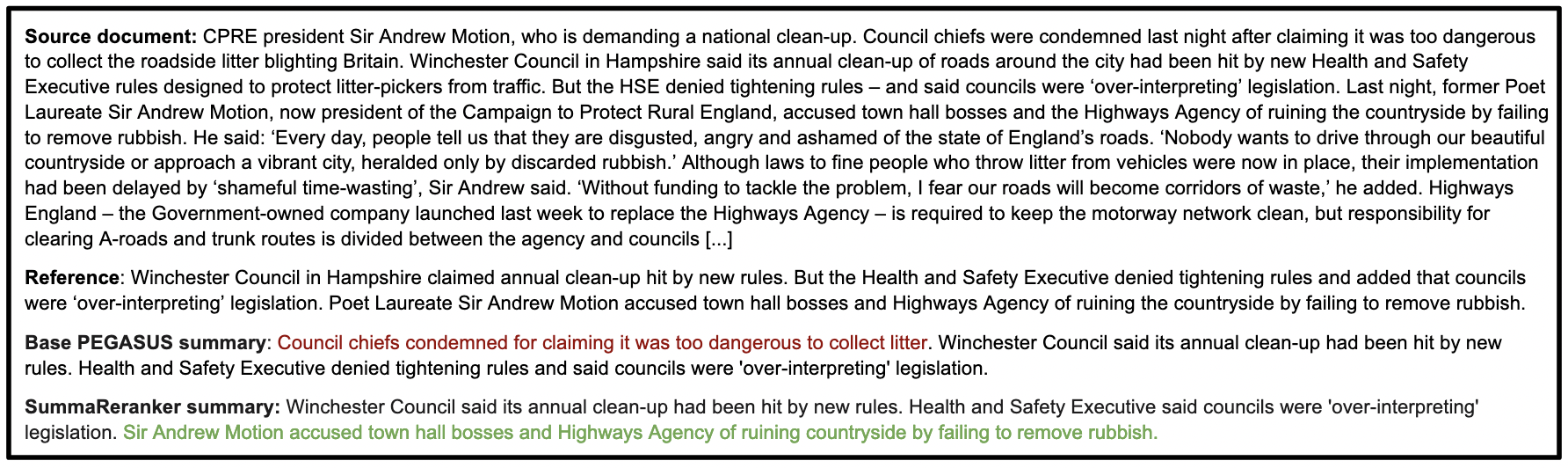}
    % \vspace{-0.5em}
    \caption{\textbf{Example of a summary} generated by SummaReranker trained for \{R-1, R-2, R-L\} on \textbf{CNN/DM}. The sentence in green is included in the SummaReranker summary, while the one in red is discarded.}
    \label{fig:4}
    % \vspace{-1.5em}
\end{figure*}

%%%%%%%%%%%%%%%%%%%

%%%%%%%%%%%%%%%%%%%
\subsection{Ranking Evaluation}

Beyond summary properties, we investigate the performance of re-ranking itself with rank-based evaluation measures. A perfect re-ranker should always single out the best summary from the rest, yielding oracle results. To evaluate how SummaReranker ranks the best summary, we compute the best summary candidate recall at different thresholds. 
% For a single best summary candidate among $m$ candidates, the recall at $k$ for a random uniform ranking baseline is simply given by:
% $R@k = \frac{k}{m}$.
{Since several candidates might get the same metric scores (\Cref{sec:appendix_c}), the best candidate recall at threshold $k$ for the random uniform ranking baseline is not the standard $R@k = \frac{k}{m}$ anymore but becomes instead}:
% \vspace{-1.0em}
\begin{equation}
\label{Eq:6}
    R@k = \frac{\binom{m}{m_{\text{best}}} - \binom{m-k}{m_{\text{best}}}}{\binom{m}{m_{\text{best}}}}
    % \vspace{-0.25em}
\end{equation}

\noindent where ${m_{\text{best}}}$ is the number of best candidates.

Following \Cref{fig:3}, a PEGASUS with diverse beam search ranking of summary candidates (dashed lines) is not significantly better than the corresponding random baseline from \cref{Eq:6} (dotted lines) on CNN/DM and Reddit TIFU. However, it improves on it on XSum, confirming the observation made in \Cref{tab:6} that it is harder to train a re-ranker on this dataset. On all three datasets, SummaReranker (solid lines) significantly pushes the recall at all thresholds. We note +14.90 absolute recall@5 improvement on CNN/DM (50.84 versus 35.94, indicated by the black arrow), +9.54 on XSum and +5.23 on Reddit TIFU.

%%%%%%%%%%%%%%%%%%%%%%%%
\subsection{Qualitative Evaluation}

Lastly, we demonstrate that re-ranking improvements in quantitative metrics also translate to qualitatively better summaries. \Cref{fig:4} shows an example of summary selected by SummaReranker, alongside its source document, ground-truth (reference) summary and output from the base model. SummaReranker is able to include a whole sentence which was missed by the base summary. We refer to \Cref{sec:appendix_k} for full re-ranking {demonstrations} on each of the three datasets.

\begin{figure}
    \centering 
    \includegraphics[width=0.99\columnwidth]{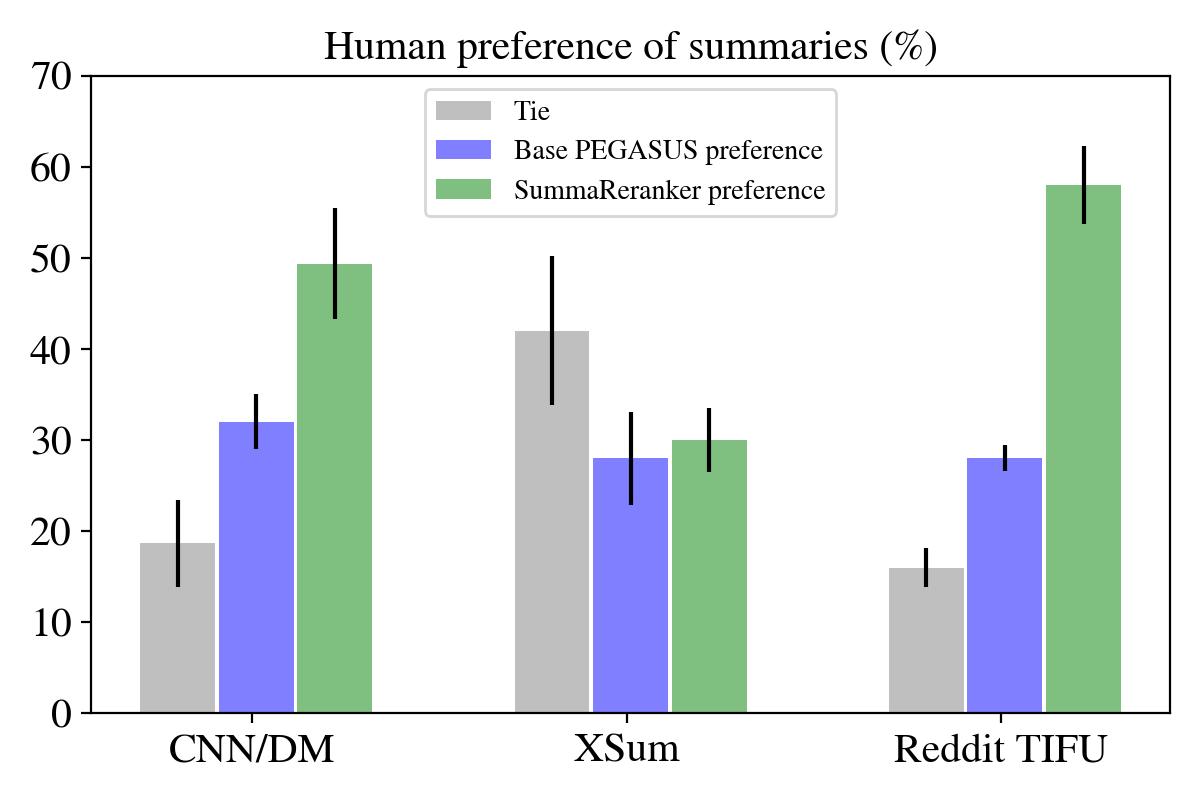}
    % \vspace{-2.0em}
    \caption{\textbf{Human evaluation} results on all three datasets. Black vertical bars are standard deviation across human raters.}
    \label{fig:5}
    % \vspace{-1.2em}
\end{figure}

We also conduct a human evaluation. We asked three different humans to evaluate 50 randomly sampled test summaries {for each dataset}. Human raters were graduate students with professional English proficiency (TOEFL scores above 100 out of 120). Humans were shown the source document alongside the top beam search summary from PEGASUS, and the corresponding summary candidate selected by SummaReranker. They were asked to choose which one they believe is more faithful. They could choose a tie, because in some cases the base summary and the re-ranked one are very similar, or even identical (\Cref{sec:appendix_i}). In \Cref{fig:5}, we see that on average, humans are more likely to pick the SummaReranker candidate. 

% \vspace{-0.3em}

%%%%%%%%% FIGURE 5: human eval %%%%%%%%%%

%%%%%%%%%%%%%%%%%%%

%% file: Sections/discussion.tex
%%%%%%%%% FIGURE 6 %%%%%%%%%%

\begin{figure}[h]
    \centering 
    \includegraphics[width=0.99\columnwidth]{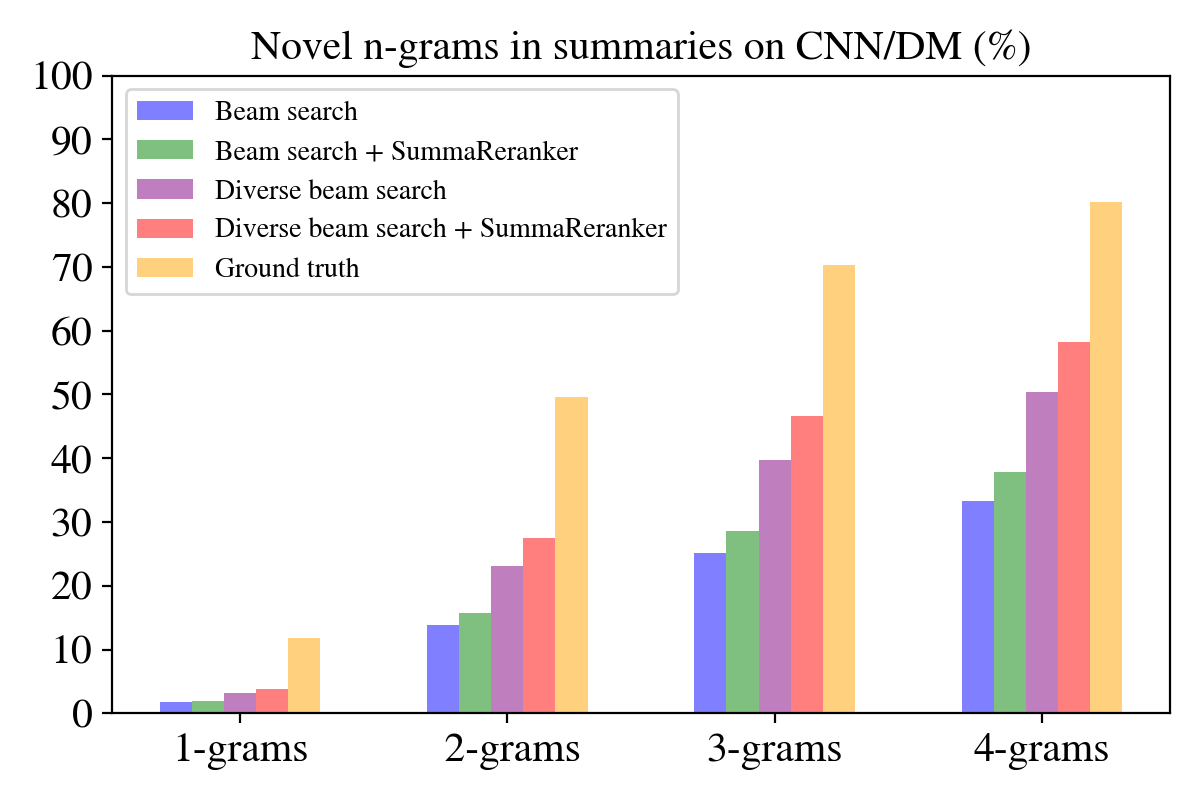}
    \includegraphics[width=0.99\columnwidth]{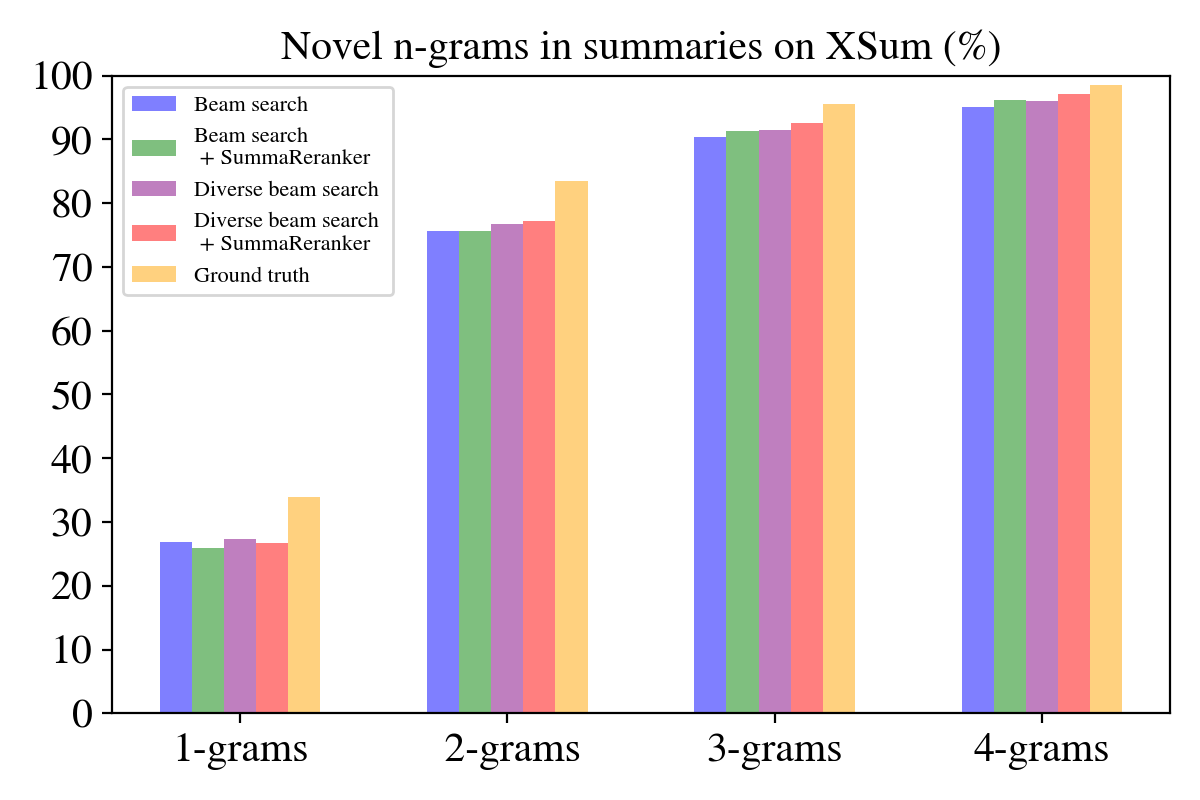}
    \includegraphics[width=0.99\columnwidth]{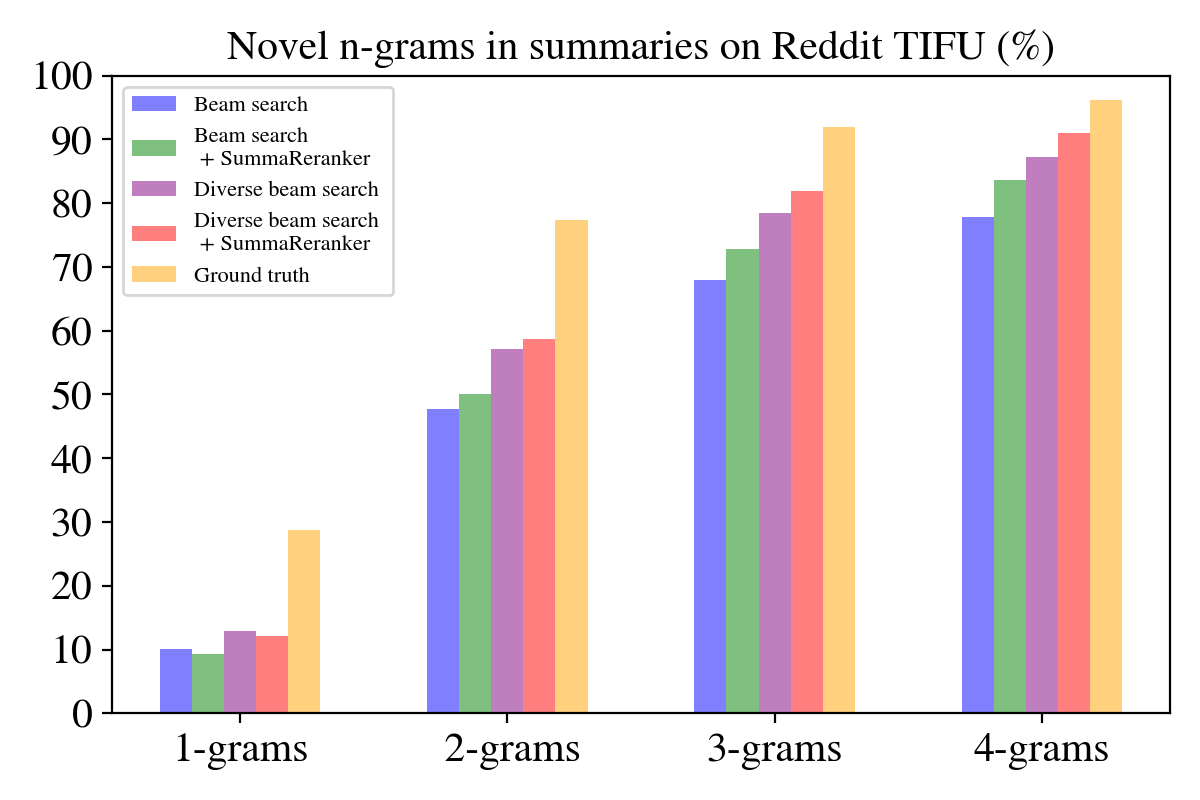}
    % \vspace{-2em}
    \caption{\textbf{Novel $n$-grams} with PEGASUS, across all datasets and with beam search and diverse beam search.}
    \label{fig:6}
    % \vspace{-1.0em}
\end{figure}

%%%%%%%%%%%%%%%%%%%

%%%%%%%%%%%%%%%%%%%%%%%%
\paragraph{Abstractiveness}

Given that we are not modifying the base model nor its training procedure, we analyze whether our re-ranking system favors more abstractive candidates. In \Cref{fig:6}, we display the percentage of novel $n$-grams for $n$ in \{1,2,3,4\}, for a base PEGASUS with beam search (blue) and diverse beam search (purple) decoding, and when adding SummaReranker in both cases (green and red, respectively). As first raised in \cite{see-etal-2017-get}, summary candidates are much less abstractive than ground truth summaries on CNN/DM. Yet, our re-ranker selects more abstractive candidates according to all $n$-grams metrics, even more so with diverse beam search, which is already more abstractive than beam search. This observation also holds on Reddit TIFU and XSum (other than 1-grams). XSum summary candidates are already almost as abstractive as the ground truth and it is harder to obtain significant abstractiveness gains through our re-ranking.

%%%%%%%%%%%%%%%%%%%%%%%%
% \vspace{-0.5em}
{\paragraph{Speed/Performance trade-off} On top of base model training and candidate generation, SummaReranker inference cost is linear in the number of candidates. A single candidate takes on average 38ms to be scored. As seen in \Cref{tab:5} and \Cref{tab:6}, the performance gains from mixing several decoding methods to generate summary candidates are not scaling consistently (all four decoding methods are not better than just beam search and diverse beam search). To provide more insights on the speed/performance trade-off, we show in \Cref{sec:appendix_j} SummaReranker performance when randomly sub-sampling $k \in \{ 1, \dots , 15 \}$ candidates. On CNN/DM, re-ranking as few as two candidates is sufficient to improve on the baseline PEGASUS. On XSum, it needs three to eight, and on Reddit TIFU three to four. As a rule of thumb, it is better to score all candidates when possible, but six to eight candidates provide a good trade-off between speed and performance across datasets.}

%%%%%%%%%%%%%%%%%%%%%%%%
\paragraph{Further Work}

% long source documents
To encode the source jointly with the summary candidate, we need to truncate the source to a fixed number of tokens. Thus, we are limited by the maximum context window of the language model encoder (512 in the case of RoBERTa-large). Applying SummaReranker to long-document summarization, such as scientific articles summarization \cite{cohan-etal-2018-discourse} would need better long-range modeling.
% a different encoder, capable of modeling a long amplitude of source-candidate interactions. 
% weighting metrics
In \cref{sec:model}, we weighted metric-dependent losses uniformly. We leave to further work the exploration of more complex weight balancing or multi-task learning objectives \cite{lin2019pareto}.

%% file: Sections/conclusion.tex
\vspace{-0.5em}
We introduced SummaReranker, the first multi-task re-ranking framework for abstractive summarization. Encoding the source with the candidate, our model predicts whether the summary candidate maximizes each of the metrics optimized for.
% Using the source, our model predicts which summary candidate maximizes each of the evaluation metrics optimized for.
SummaReranker works well across diverse datasets, models, decoding methods and summarization evaluation metrics. Summaries selected by SummaReranker improve the ROUGE state-of-the-art on CNN/DM and XSum. In addition, we also show that they are more abstractive and more likely to be preferred by human evaluators over base model outputs.

% {\color{red}We introduced SummaReranker, a system which ranks summary candidates using the source. [Is this the topic sentence for your Conclusion? Is this the key differentiator of your technical novelty?]} SummaReranker works well over different domains (news and social media), datasets (CNN/DM, XSum, Reddit TIFU), base models (BART, PEGASUS), decoding methods (beam search, diverse beam search, top-k sampling, top-p sampling), and summarization metrics (ROUGE, BERTScore, BARTScore). 
% {\color{blue}Summaries selected by SummaReranker improve performance of the metrics optimized for over the base summaries, which includes pushing the ROUGE state-of-the-art on CNN/DM and XSum. [Suggest to rewrite this sentence. Hard to understand. Maybe break down to two sentences, where ROUGE SOTA is a separate sentence. But even then, the first clause can be worded better]} {\color{red}} 

%% file: Sections/acknowledgements.tex
This research was supported by the SINGA scholarship and partially supported by the National Research Foundation, Prime Minister’s Office, Singapore under its Campus for Research Excellence and Technological Enterprise (CREATE) programme. We would like to thank anonymous reviewers for their insightful feedback on how to improve the paper.

%% file: Sections/appendix_a.tex
For evaluation metrics, we used the following packages:
\begin{itemize}
    \item For ROUGE metrics \cite{lin-hovy-2003-automatic}, we used the public \emph{rouge-score} package from Google Research: \\ \url{https://github.com/google-research/google-research/tree/master/rouge}
    \item For BERTScore \cite{zhang2019bertscore}, we used the public \emph{bert-score} package shared by the authors:\\
    \url{https://github.com/Tiiiger/bert_score}
    \item For BARTScore \cite{yuan2021bartscore}, we used the public code shared by the authors:\\
    \url{https://github.com/neulab/BARTScore}
\end{itemize}

\begin{table}[h]
\centering
\resizebox{\columnwidth}{!}{
\begin{tabular}{cccccccccc}
\toprule
\textbf{Dataset}               
& \textbf{Model} 
& \textbf{LR} 
& \textbf{Epochs} 
& \textbf{Opt.} 
& \textbf{BS} 
& \textbf{LS} 
& \textbf{MP} 
& \textbf{\begin{tabular}[c]{@{}c@{}}Source\\ tokens\end{tabular}} 
& \textbf{\begin{tabular}[c]{@{}c@{}}Summary \\ tokens\end{tabular}} \\
\midrule
\multirow{2}{*}{CNN/DM} & PEGASUS  & 5e-5 & 10 & Adafactor & 256 & 0.1 & No & 1024 & 128 \\
                               & BART & 3e-5 & 10 & Adam & 80 & 0.1 & Yes & 1024 & 128 \\
\hdashline
\multirow{2}{*}{XSum}          & PEGASUS & 5e-5 & 10 & Adafactor & 256 & 0.1 & No & 512 & 64  \\
                               & BART & 3e-5 & 10 & Adam & 80 & 0.1 & Yes & 512 & 64 \\
\hdashline
\multirow{2}{*}{Reddit TIFU}   & PEGASUS & 1e-4 & 15 & Adafactor & 256 & 0.1 & No & 512 & 128 \\
                               & BART & 3e-5 & 15 & Adam & 80& 0.1 & Yes & 512 & 128      \\
\bottomrule
\end{tabular}
}
\caption{\textbf{Hyper-parameters} for \textbf{fine-tuning} the base models. \textbf{LR} designates the \emph{learning rate}, \textbf{Epochs} is the number of epochs, \textbf{Opt.} is the \emph{optimizer}, \textbf{BS} is the \emph{batch size}, \textbf{LS} means \emph{label smoothing}, and \textbf{MP} means \emph{mixed precision}. \textbf{Source tokens} is the maximum size of the input document, \textbf{Summary tokens} the maximum size of the output summary.}
\label{tab:7}
\end{table}

\begin{table}[h]
\centering
\resizebox{\columnwidth}{!}{
\begin{tabular}{ccccccc}
\toprule
\textbf{Dataset}               
& \textbf{Model} 
& \textbf{\begin{tabular}[c]{@{}c@{}}Source\\ tokens\end{tabular}} 
& \textbf{\begin{tabular}[c]{@{}c@{}}Summary \\ tokens\end{tabular}} 
& \textbf{\begin{tabular}[c]{@{}c@{}}Length\\ penalty\end{tabular}} 
& \textbf{\begin{tabular}[c]{@{}c@{}}Repetition\\ penalty\end{tabular}} 
& \textbf{\begin{tabular}[c]{@{}c@{}}Trigram\\ blocking\end{tabular}} \\
\midrule
\multirow{2}{*}{CNN/DM} & PEGASUS & 1024 & 128 & 0.8 & 1.0  & No \\
                               & BART & 1024 & 128 & 0.8 & 1.0 & No \\
\hdashline
\multirow{2}{*}{XSum}          & PEGASUS & 512 & 64 & 0.8 & 1.0 & Yes \\
                               & BART & 512 & 64 & 0.8 & 1.0 & Yes \\
\hdashline
\multirow{2}{*}{Reddit TIFU}   & PEGASUS & 512 & 128 & 0.6 & 1.0 & Yes \\
                               & BART & 512 & 128 & 1.0  & 1.0  & Yes \\
\bottomrule
\end{tabular}
}
\caption{\textbf{Hyper-parameters} for the summary candidates \textbf{generation} with the base models.}
\label{tab:8}
\end{table}

%% file: Sections/appendix_b.tex
\def \hfillx {\hspace*{-\textwidth} \hfill}

\begin{table}[h]
  \centering
  \resizebox{\columnwidth}{!}{
    \begin{tabular}{lcccccc}
    % \textbf{\begin{tabular}[c]{@{}c@{}}Decoding\\ methods\end{tabular}} &
    \toprule
    \textbf{Decoding methods} &
    \textbf{\begin{tabular}[c]{@{}c@{}}\# Summary\\ candidates\end{tabular}} & \textbf{R-1} & \textbf{R-2} & \textbf{R-L} & \textbf{BS} & \textbf{BaS} \\
    \midrule
    Beam search (top beam) & 1 & 47.33 & 24.75 & 39.43 & 92.01 & -1.92\\
    \midrule
    Beam search & 15  & 56.07  & 33.80  & 48.33 & 93.19 & -1.82  \\
    Diverse beam search & 15 & \textbf{57.82} & \textbf{35.28} & \textbf{50.95} & \textbf{93.65} & \textbf{-1.63} \\
    Top-k sampling & 15 & 55.57 & 32.54 & 48.35 & 93.18 & -1.86 \\
    Top-p sampling & 15 & 56.74 & 33.94 & 49.60 & 93.40 & -1.77 \\
    \hdashline
    % Both  & 30 & \textbf{60.07} & \textbf{38.19} & \textbf{53.40} & \textbf{93.94} & \textbf{-1.52} \\
    All four above & 60 & \textbf{62.30} & \textbf{40.84} & \textbf{55.92} & \textbf{94.24} & \textbf{-1.48} \\
    \bottomrule
    \end{tabular}
    }
    \caption{\textbf{Oracle scores} for four popular decoding methods and five summarization evaluation measures for a base PEGASUS model on \textbf{XSum}.}
    \label{tab:9}
\end{table}

\begin{table}[h]
  \centering
  \resizebox{\columnwidth}{!}{
    \begin{tabular}{lcccccc}
    % \textbf{\begin{tabular}[c]{@{}c@{}}Decoding\\ methods\end{tabular}} &
    \toprule
    \textbf{Decoding methods} &
    \textbf{\begin{tabular}[c]{@{}c@{}} \# Summary\\ candidates\end{tabular}} & \textbf{R-1} & \textbf{R-2} & \textbf{R-L} & \textbf{BS} & \textbf{BaS} \\
    \midrule
    Beam search (top beam) & 1 & 26.28& 9.01 & 21.52 & 87.34 & -3.46  \\
    \midrule
    Beam search & 15 & 36.08 & 14.93 & 29.70 & 88.64 & -2.89 \\
    Diverse beam search & 15 & 36.70 & 15.22 & \textbf{30.88} & \textbf{89.08} & \textbf{-2.81} \\
    Top-k sampling & 15 & 36.76 & 14.37 & 29.49 & 88.53 & -3.14 \\
    Top-p sampling & 15 & \textbf{37.54} & \textbf{15.24} & 30.50 & 88.69 & -3.03 \\
    \hdashline
    % Both & 30 & \textbf{40.00} & \textbf{18.04} & \textbf{33.70} & \textbf{89.42} & \textbf{-2.65} \\
    All four above & 60 & \textbf{43.25} & \textbf{20.70} & \textbf{36.41} & \textbf{89.71} & \textbf{-2.58} \\
    \bottomrule
    \end{tabular}
    }
    \caption{\textbf{Oracle scores} for four popular decoding methods and five summarization evaluation measures for a base PEGASUS model on \textbf{Reddit TIFU}.}
    \label{tab:10}
\end{table}

Observations from \Cref{tab:9} and \Cref{tab:10} are consistent with the ones made in \Cref{tab:1}: oracle scores are widely above the top beam baseline, and keep increasing when mixing several decoding methods.

%% file: Sections/appendix_c.tex
% Please add the following required packages to your document preamble:
% \usepackage{multirow}
\begin{table}[h]
\centering
\resizebox{0.95\columnwidth}{!}{
\begin{tabular}{cccccccc}
\toprule
\textbf{} & \textbf{} & \textbf{} & \multicolumn{5}{c}{\textbf{Scoring metric}} \\
\textbf{Dataset} & \textbf{Model} & \textbf{\begin{tabular}[c]{@{}c@{}}Generation\\ method\end{tabular}} & \textbf{R-1} & \textbf{R-2} & \textbf{R-L} & \textbf{BS} & \textbf{BaS} \\
\midrule
\multirow{6}{*}{CNN/DM} & \multirow{4}{*}{PEGASUS} & \{1\} & 11.51 & 10.87 & 11.54 & 14.96 & 14.96 \\
                        &                          & \{2\} & 14.34 & 14.09 & 14.34 & 14.99 & 14.99 \\
                        &                          & \{3\} & 14.65 & 14.40 & 14.65 & 14.99 & 14.99 \\
                        &                          & \{4\} & 14.68 & 14.41 & 14.69 & 15.00 & 15.00 \\
\cdashline{2-8} & \multirow{2}{*}{BART} & \{1\} & 11.51 & 10.90 & 11.54 & 14.93 & 14.95 \\
                &                       & \{2\} & 13.89 & 13.71 & 13.89 & 14.80 & 14.79 \\
\midrule
\multirow{4}{*}{XSum} & \multirow{2}{*}{PEGASUS} & \{1\} & 8.90 & 7.91 & 8.56 & 14.99 & 14.99 \\
                      &                          & \{2\}& 12.05 & 10.92 & 12.11 & 14.97 & 14.98 \\
\cdashline{2-8} & \multirow{2}{*}{BART} & \{1\} & 8.70 & 7.57 & 8.33 & 14.99 & 15.00 \\
                &                       & \{2\} & 7.37 & 6.63 & 7.37 & 14.59 & 14.99 \\
\midrule
\multirow{4}{*}{Reddit TIFU} & \multirow{2}{*}{PEGASUS} & \{1\} & 9.19 & 6.31 & 8.85 & 14.99 & 14.99 \\
                             &                          & \{2\} & 7.84 & 5.06 & 7.77 & 14.89 & 14.97 \\
\cdashline{2-8} & \multirow{2}{*}{BART} & \{1\} & 7.73 & 5.15 & 7.56 & 14.99 & 14.99 \\
                &                       & \{2\} & 7.42 & 3.92 & 7.38 & 14.89 & 14.97 \\
\bottomrule
\end{tabular}
}
\caption{\textbf{Number of unique scores} among pools of 15 candidates generated on different datasets (CNN/DM, XSum, Reddit TIFU) with different base models (PEGASUS, BART) and different decoding methods (\{1\} stands for beam search, \{2\} is diverse beam search, \{3\} is top-p sampling and \{4\} top-k sampling). The lowest possible score of 1 indicates that all 15 candidates are assigned the same score under the metric being considered, while the highest of 15 means that all candidates are assigned a different score.}
\label{tab:11}
\end{table}

In \Cref{tab:11}, BERTScore (BS) and BARTScore (BaS) have results closer to 15, indicating that it is unlikely that two summary candidates share the exact metric score. This is understandable given that both these metrics are based on embeddings from pre-trained language models (BERT and BART, respectively), and embeddings values will vary whenever the input text is different, making it unlikely to have two candidates collude on the same score. In contrast, ROUGE measures n-gram overlaps, and two different summary candidates might get the same ROUGE score with the target summary (for instance if they only differ by n-grams not present in the target).

%% file: Sections/appendix_d.tex
% Please add the following required packages to your document preamble:
% \usepackage{multirow}
\begin{table}[h]
\centering
\resizebox{0.95\columnwidth}{!}{
\begin{tabular}{cccccccc}
\toprule
\textbf{} & \textbf{} & \textbf{} & \multicolumn{5}{c}{\textbf{Scoring metric}} \\
\textbf{Dataset} & \textbf{Model} & \textbf{\begin{tabular}[c]{@{}c@{}}Generation\\ method\end{tabular}} & \textbf{R-1} & \textbf{R-2} & \textbf{R-L} & \textbf{BS} & \textbf{BaS} \\
\midrule
\multirow{6}{*}{CNN/DM} & \multirow{4}{*}{PEGASUS} & \{1\} & 0.00 & 0.50 & 0.00 & 0.00 & 0.00 \\
                        &                          & \{2\} & 0.00 & 0.03 & 0.00 & 0.00 & 0.00 \\
                        &                          & \{3\} & 0.03 & 0.06 & 0.03 & 0.03 & 0.00 \\
                        &                          & \{4\} & 0.00 & 0.03 & 0.00 & 0.00 & 0.00 \\
\cdashline{2-8} & \multirow{2}{*}{BART} & \{1\} & 0.00 & 0.47 & 0.00 & 0.00 & 0.00 \\
                &                       & \{2\} & 0.00 & 0.05 & 0.00 & 0.00 & 0.00 \\
\midrule
\multirow{4}{*}{XSum} & \multirow{2}{*}{PEGASUS} & \{1\} & 0.06 & 3.34 & 0.10 & 0.00 & 0.00 \\
                      &                          & \{2\} & 0.04 & 1.11 & 0.04 & 0.00 & 0.00 \\
\cdashline{2-8} & \multirow{2}{*}{BART} & \{1\} & 0.17 & 4.31 & 0.19 & 0.00 & 0.00 \\
                &                       & \{2\} & 0.04 & 2.59 & 0.04 & 0.00 & 0.00 \\
\midrule
\multirow{4}{*}{Reddit TIFU} & \multirow{2}{*}{PEGASUS} & \{1\} & 2.04 & 21.15 & 2.04 & 0.00 & 0.00 \\
                             &                          & \{2\} & 1.52 & 17.03 & 1.52 & 0.00 & 0.00 \\
\cdashline{2-8} & \multirow{2}{*}{BART} & \{1\} & 2.32 & 24.14 & 2.32 & 0.00 & 0.00 \\
                &                       & \{2\} & 1.73 & 21.60 & 1.73 & 0.00 & 0.00 \\
\bottomrule
\end{tabular}
}
\caption{\textbf{Fraction of sets of candidates with all identical scores (\%)} for pools of 15 candidates generated on different datasets (CNN/DM, XSum, Reddit TIFU) with different base models (PEGASUS, BART) and different decoding methods (\{1\} stands for beam search, \{2\} is diverse beam search, \{3\} is top-p sampling and \{4\} top-k sampling. 
}
\label{tab:12}
\end{table}

We note that cases where all scores are identical are a small minority. ROUGE-2 is more likely than other metrics to lead to such a scenario of all identical scores.

%% file: Sections/appendix_e.tex
\begin{table}[h]
\renewcommand{\arraystretch}{0}
\setlength{\fboxsep}{3mm} % box size
\setlength{\tabcolsep}{0pt}
  \centering
  \resizebox{\columnwidth}{!}{
    \begin{tabular}{c*{5}{R}}
      & \textbf{R-1} & \textbf{R-2} & \textbf{R-L} & \textbf{BS} & \textbf{BaS} \EndTableHeader\\
      \noalign{\vskip 1mm} 
      \textbf{R-1}~~ & 1.000 & 0.888 & 0.905 & 0.850 & 0.657 \\
      \textbf{R-2}~~ & 0.888 & 1.000 & 0.911 & 0.790 & 0.628 \\
      \textbf{R-L}~~ & 0.905 & 0.911 & 1.000 & 0.847 & 0.620 \\
      \textbf{BS}~~  & 0.850 & 0.790 & 0.847 & 1.000 & 0.690 \\
      \textbf{BaS}~~ & 0.657 & 0.628 & 0.620 & 0.690 & 1.000 \\
    \end{tabular}
    }
    \caption{\textbf{Pearson correlation coefficient} between the five evaluation metrics \{R-1, R-2, R-L, BS, BaS\} for a base PEGASUS decoded with beam search on \textbf{XSum}.}
    \label{tab:13}
\end{table}

\begin{table}[h]
\renewcommand{\arraystretch}{0}
\setlength{\fboxsep}{3mm} % box size
\setlength{\tabcolsep}{0pt}
  \centering
  \resizebox{\columnwidth}{!}{
    \begin{tabular}{c*{5}{R}}
      & \textbf{R-1} & \textbf{R-2} & \textbf{R-L} & \textbf{BS} & \textbf{BaS} \EndTableHeader\\
      \noalign{\vskip 1mm} 
      \textbf{R-1}~~ & 1.000 & 0.806 & 0.927 & 0.766 & 0.600 \\
      \textbf{R-2}~~ & 0.806 & 1.000 & 0.856 & 0.679 & 0.524 \\
      \textbf{R-L}~~ & 0.927 & 0.856 & 1.000 & 0.768 & 0.564 \\
      \textbf{BS}~~  & 0.766 & 0.679 & 0.768 & 1.000 & 0.646 \\
      \textbf{BaS}~~ & 0.600 & 0.524 & 0.564 & 0.656 & 1.000 \\
    \end{tabular}
    }
    \caption{\textbf{Pearson correlation coefficient} between the five evaluation metrics \{R-1, R-2, R-L, BS, BaS\} for a base PEGASUS decoded with beam search on \textbf{Reddit TIFU}.}
    \label{tab:14}
\end{table}

Metrics correlation from \Cref{tab:13} and \Cref{tab:14} follow the same pattern as in \Cref{tab:2}.

%% file: Sections/appendix_f.tex
\begin{table}[h]
    \centering 
    \resizebox{\columnwidth}{!}{
    \begin{tabular}{lcccccc}
    \toprule
    \textbf{Model} 
    & \textbf{\begin{tabular}[c]{@{}c@{}}Model\\ stage\end{tabular}} 
    & \textbf{\begin{tabular}[c]{@{}c@{}}Decoding\\ methods ($\sD$)\end{tabular}} 
    & \textbf{R-1} 
    & \textbf{R-2} 
    & \textbf{R-L} 
    & \textbf{\begin{tabular}[c]{@{}c@{}}Gain\\ (\%)\end{tabular}} 
    \\
    \midrule
    % \rowcolor{gray!20} 
    PEGASUS - 1st half & 1 & \{1\} & 46.02 & 23.38 & 38.10 & \_ \\
    PEGASUS - 1st half & 1 & \{2\} & 45.41 & 22.37 & 37.22 & \_\\
    % \rowcolor{gray!20} 
    PEGASUS - 2nd half & 1 & \{1\} & \textbf{46.26} & \textbf{23.45} & \textbf{38.22} & \_ \\
    PEGASUS - 2nd half & 1 & \{2\} & 45.53 & 22.42 & 37.31 & \_ \\
    \hdashline
    % \rowcolor{gray!20} 
    BART - 1st half & 1 & \{1\} & 42.76 & 20.22 & 35.00 & \_ \\
    BART - 1st half & 1 & \{2\} & 40.93	& 18.75	& 33.44 & \_ \\
    % \rowcolor{gray!20} 
    BART - 2nd half & 1 & \{1\} & 42.63	& 20.22	& 35.08 & \_ \\
    BART - 2nd half & 1 & \{2\} & 40.65	& 18.65	& 33.38 & \_ \\
    \midrule
    % \rowcolor{gray!20} 
    PEGASUS - 1st half + SR & 2 & \{1\} & 45.01 & 22.06 & 36.92 & -3.63 \\
    PEGASUS - 1st half + SR & 2 & \{2\} & \textbf{46.35} & \textbf{22.64} & \textbf{38.05} & 0.83 \\
    % \rowcolor{gray!20} 
    PEGASUS - 2nd half + SR & 2 & \{1\} & 45.25 & 22.10 & 36.96 & -3.77 \\
    PEGASUS - 2nd half + SR & 2 & \{2\} & 46.25 & 22.50 & 37.93 & 1.17 \\
    \hdashline
    % \rowcolor{gray!20} 
    BART - 1st half + SR & 2 & \{1\} & 44.09 & 20.71 & 35.86 & 2.67 \\
    BART - 1st half + SR & 2 & \{2\} & 43.70 & 19.90 & 35.21 & 6.07 \\
    % \rowcolor{gray!20} 
    BART - 2nd half + SR & 2 & \{1\} & 44.30 & 20.88 & 36.23 & 3.50 \\
    BART - 2nd half + SR & 2 & \{2\} & 43.96 & 20.03 & 35.49 & 7.27 \\
    \midrule
    PEGASUS - 1st half + SR & 2 & \{1, 2\} & 46.74 & 23.10 & 38.35 & 0.37 \\
    PEGASUS - 2nd half + SR & 2 & \{1, 2\} & \textbf{47.00} & \textbf{23.30} & \textbf{38.54} & 0.60 \\
    \hdashline
    BART - 1st half + SR & 2 & \{1, 2\} & 44.52 & 20.59 & 35.93 & 2.87 \\
    BART - 2nd half + SR & 2 & \{1, 2\} & 44.68	& 20.76	& 36.20 & 3.57 \\
    \bottomrule
    \end{tabular}
    }
    \caption{\textbf{Base setup results} for SummaReranker applied to PEGASUS and BART on the \textbf{XSum} dataset. \textbf{SR} refers to SummaReranker. \textbf{Decoding method \{1\} is beam search, \textbf{\{2\}} is diverse beam search}. Best scores for each type of model are in bold. \textbf{Gain} represents the mean relative gain over \{R-1, R-2, R-L\} compared to the best decoding method.}
    \label{tab:15}
\end{table}

\begin{table}[h]
    \centering 
    \resizebox{\columnwidth}{!}{
    \begin{tabular}{lcccccc}
    \toprule
    \textbf{Model} 
    & \textbf{\begin{tabular}[c]{@{}c@{}}Model\\ stage\end{tabular}} 
    & \textbf{\begin{tabular}[c]{@{}c@{}}Decoding\\ methods ($\sD$)\end{tabular}} 
    & \textbf{R-1} 
    & \textbf{R-2} 
    & \textbf{R-L} 
    & \textbf{\begin{tabular}[c]{@{}c@{}}Gain\\ (\%)\end{tabular}} 
    \\
    \midrule
    % \rowcolor{gray!20} 
    PEGASUS - 1st half & 1 & \{1\} & 24.83 & 8.29 & 20.38 & \_ \\
    PEGASUS - 1st half & 1 & \{2\} & 23.77 & 7.38 & 19.37 & \_\\
    % \rowcolor{gray!20} 
    PEGASUS - 2nd half & 1 & \{1\} & 25.16 & 8.42 & 20.53 & \_ \\
    PEGASUS - 2nd half & 1 & \{2\} & 24.18 & 7.53 & 19.68 & \_ \\
    \hdashline
    % \rowcolor{gray!20} 
    BART - 1st half & 1 & \{1\} & 28.38 & \textbf{9.60} & 22.44 & \_ \\
    BART - 1st half & 1 & \{2\} & \textbf{28.60} & 8.96 & \textbf{22.49} & \_ \\
    % \rowcolor{gray!20} 
    BART - 2nd half & 1 & \{1\} & 26.94	& 9.13 & 21.65 & \_ \\
    BART - 2nd half & 1 & \{2\} & 25.83	& 8.38 & 20.97 & \_ \\
    \midrule
    % \rowcolor{gray!20} 
    PEGASUS - 1st half + SR & 2 & \{1\} & 28.78 & 9.20 & 22.74 & 12.83 \\
    PEGASUS - 1st half + SR & 2 & \{2\} & 28.63	& 8.71 & 22.71 & 18.53 \\
    % \rowcolor{gray!20} 
    PEGASUS - 2nd half + SR & 2 & \{1\} & 28.87	& 9.24 & 22.73 & 11.70 \\
    PEGASUS - 2nd half + SR & 2 & \{2\} & 28.41	& 8.46 & 22.44 & 14.63 \\
    \hdashline
    % \rowcolor{gray!20} 
    BART - 1st half + SR & 2 & \{1\} & 28.98 & \textbf{9.62} & \textbf{22.96} & 1.53 \\
    BART - 1st half + SR & 2 & \{2\} & \textbf{28.89} & 8.70 & 22.40 & -0.77 \\
    % \rowcolor{gray!20} 
    BART - 2nd half + SR & 2 & \{1\} & 27.93 & 9.48 & 22.45 & 3.73 \\
    BART - 2nd half + SR & 2 & \{2\} & 28.24 & 8.77 & 22.43 & 6.98 \\
    \midrule
    PEGASUS - 1st half + SR & 2 & \{1, 2\} & \textbf{29.93} & \textbf{9.40} & \textbf{23.50} & 16.37 \\
    PEGASUS - 2nd half + SR & 2 & \{1, 2\} & 29.65 & 9.24 & 23.22 & 13.53 \\
    \hdashline
    BART - 1st half + SR & 2 & \{1, 2\} & 29.00 & 8.78 & 22.32 & -2.63 \\
    BART - 2nd half + SR & 2 & \{1, 2\} & 29.15 & 9.11 & 22.96 & 4.70 \\
    \bottomrule
    \end{tabular}
    }
    \caption{\textbf{Base setup results} for SummaReranker applied to PEGASUS and BART on the \textbf{Reddit TIFU} dataset.}
    \label{tab:16}
\end{table}

Tables \Cref{tab:15} and \Cref{tab:16} complement the base setup results exposed in \Cref{tab:4}.

%% file: Sections/appendix_g.tex
\begin{table}[h]
\centering
\resizebox{0.95\columnwidth}{!}{
\begin{tabular}{lccccc}
\toprule
\textbf{Threshold k}                    
& \textbf{k=1} 
& \textbf{k=2} 
& \textbf{k=3} 
& \textbf{k=4} 
& \textbf{k=5} \\
\midrule
CNN-DailyMail - Random baseline         & 6.75 & 13.49 & 20.20 & 26.91 & 33.60 \\
CNN-DailyMail - PEGASUS                 & 8.57 & 15.93 & 22.76 & 29.43 & 35.94 \\
CNN-DailyMail - PEGASUS + SR & 14.97 & 25.40 & 35.00 & 43.46 & 50.84 \\
\hdashline
XSum - Random baseline                  & 8.05 & 15.81 & 23.33 & 30.62 & 37.72 \\
XSum - PEGASUS                          & 14.60 & 24.40 & 32.70 & 40.23 & 47.17 \\
XSum - PEGASUS + SR          & 16.57 & 28.60 & 39.53 & 48.78 & 56.71 \\
\hdashline
Reddit TIFU - Random baseline           & 11.39 & 21.22 & 30.35 & 38.83 & 46.70 \\
Reddit TIFU - PEGASUS                   & 14.54 & 24.11 & 33.16 & 40.10 & 48.11 \\
Reddit TIFU - PEGASUS + SR   & 16.70 & 27.07 & 37.42 & 46.02 & 53.34 \\
\bottomrule
\end{tabular}
}
\caption{Values of \textbf{recall} curves plotted in \Cref{fig:3}.}
\label{tab:17}
\end{table}

%% file: Sections/appendix_h.tex
\begin{table}[h]
\centering
\resizebox{0.95\columnwidth}{!}{
\begin{tabular}{lcc|cc|cc}
\toprule
& \multicolumn{2}{c}{\textbf{Tie}} 
& \multicolumn{2}{c}{\textbf{Base model}} 
& \multicolumn{2}{c}{\textbf{SummaReranker}} 
\\
& Mean            
& Std            
& Mean               
& Std                
& Mean                 
& Std                 
\\
\midrule
CNN/DM & 18.67 & 9.50 & 32.00 & 6.00 & 49.33 & 12.20 \\
XSum   & 42.00 & 16.33 & 28.00 & 10.20 & 30.00 & 7.12 \\
Reddit TIFU & 16.00 & 4.32 & 28.00 & 2.82 & 58.00 & 4.32 \\
\bottomrule
\end{tabular}
}
\caption{Numbers of the \textbf{human evaluation} in \Cref{fig:5}.}
\label{tab:18}
\end{table}

%% file: Sections/appendix_i.tex
\begin{table}[h]
\resizebox{0.95\columnwidth}{!}{
\begin{tabular}{ccccc}
\toprule
\textbf{Dataset}     
& \textbf{Model}   
& \begin{tabular}[c]{@{}c@{}}\textbf{Generation}\\ \textbf{method}\end{tabular} 
& \begin{tabular}[c]{@{}c@{}}\textbf{SR pick} \\ {the base}\\ candidate (\%)\end{tabular} 
& \begin{tabular}[c]{@{}c@{}}\textbf{SR pick} \\ {the best}\\ candidate (\%)\end{tabular} \\
\midrule
\multirow{4}{*}{CNN/DM} & \multirow{2}{*}{PEGASUS} & \{1\} & 3.57 & 14.81 \\
            &         & \{2\} & 11.11 & 15.00 \\
\cdashline{2-5}
            & \multirow{2}{*}{BART}    & \{1\} & \textbf{2.75} & \textbf{15.51} \\
            &         & \{2\} & 6.67 & 13.54 \\ 
\midrule
\multirow{4}{*}{XSum} & \multirow{2}{*}{PEGASUS} & \{1\} & \textbf{4.86} & 9.97 \\
            &         & \{2\} & 20.73 & 16.57 \\
\cdashline{2-5}
            & \multirow{2}{*}{BART} & \{1\} & 8.01 & 18.19 \\
            &         & \{2\} & 22.23 & \textbf{23.80} \\
\midrule
\multirow{4}{*}{Reddit TIFU} & \multirow{2}{*}{PEGASUS} & \{1\} & 6.16 & 18.21 \\
            &         & \{2\} & 16.82 & 23.09 \\
\cdashline{2-5}
            & \multirow{2}{*}{BART} & \{1\} & \textbf{3.22} & 24.04 \\
            &         & \{2\} & 3.32 & \textbf{32.88} \\                                                                            
\bottomrule
\end{tabular}
}
\caption{\textbf{Re-ranking overlap with base and best candidates.} Fraction of time that the re-ranked summary coincides with the base model one (left), and one of the best ones (oracle scores) among generated candidates (right). \textbf{SR} is SummaReranker.}
\label{tab:19}
\end{table}

In \Cref{tab:19}, we observe that SummaReranker is more likely to stick to the base model candidate with diverse beam search. Results in bold represent the most ideal scenario: SummaReranker differs the most from the base setup (lowest scores of the left column), and matches the most one of the best candidates (highest scores of the right column).

%% file: Sections/appendix_j.tex
\begin{figure}[h]
\begin{center}
\includegraphics[width=.4\textwidth]{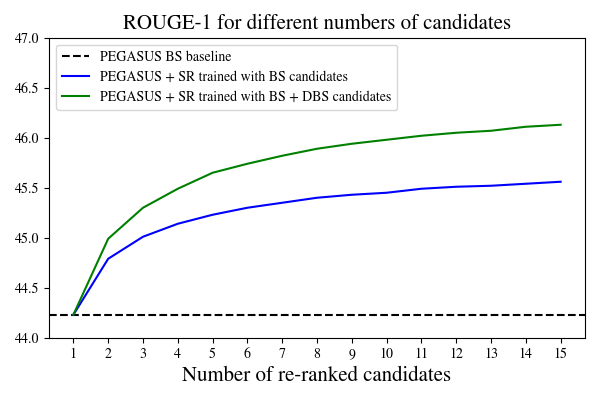}
\includegraphics[width=.4\textwidth]{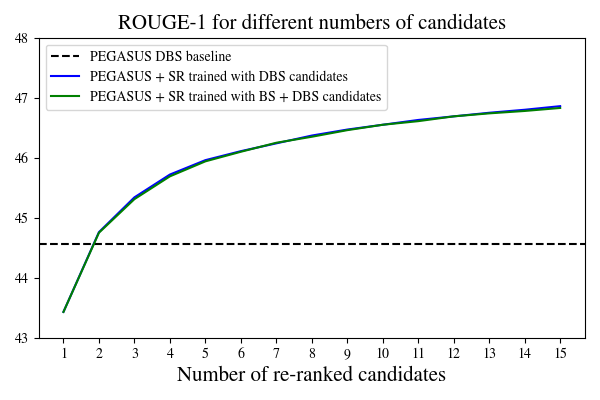}
\includegraphics[width=.4\textwidth]{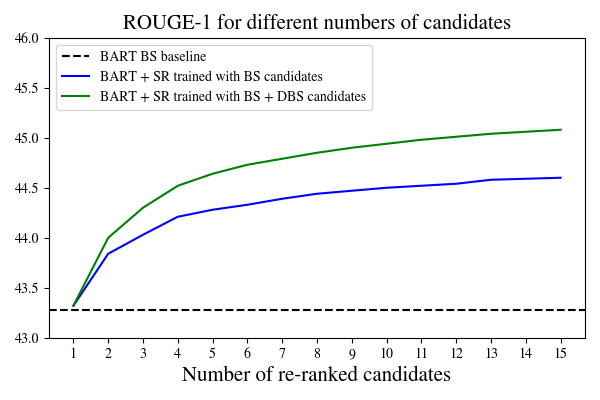}
\includegraphics[width=.4\textwidth]{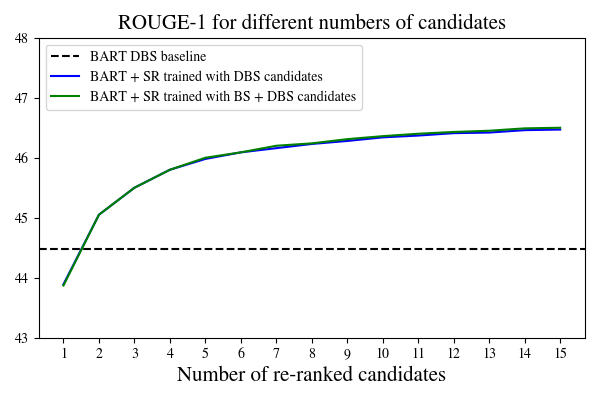}
\end{center}
\caption{\textbf{ROUGE-1 on CNN/DM} for k sampled candidates at inference time, with $k \in \{ 1, \dots , 15 \}$. \textbf{SR} stands for SummaReranker, \textbf{BS} and \textbf{DBS} refer to beam search and diverse beam search, respectively.
}
\label{fig:7}
\end{figure}

\begin{figure}
\begin{center}
\includegraphics[width=.4\textwidth]{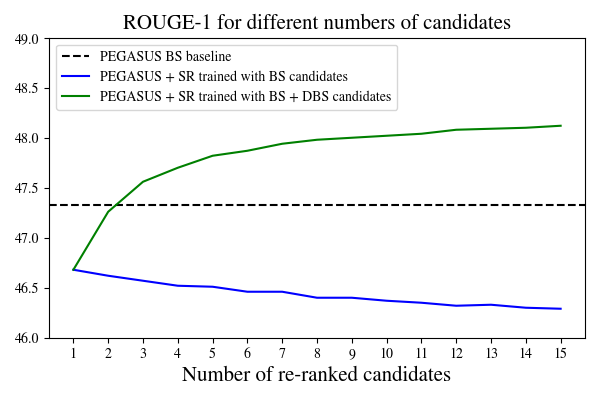}
\includegraphics[width=.4\textwidth]{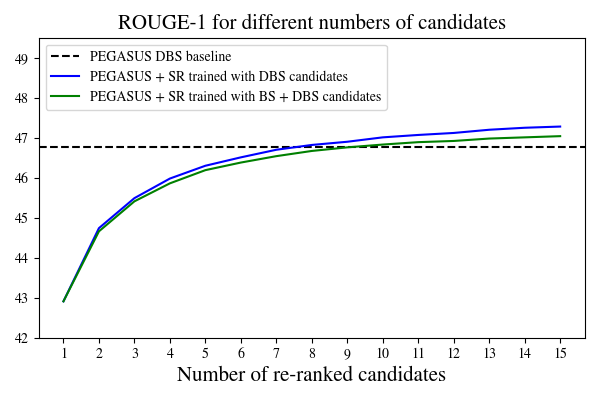}
\includegraphics[width=.4\textwidth]{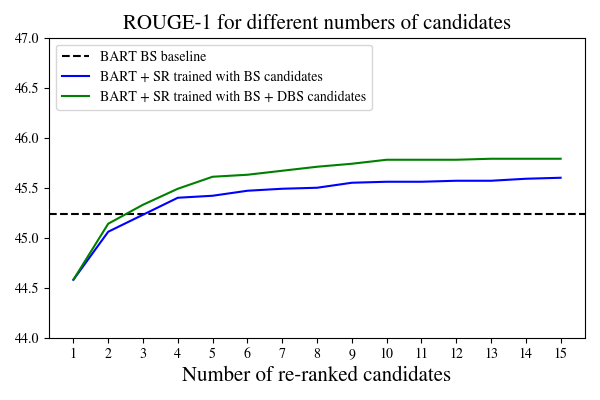}
\includegraphics[width=.4\textwidth]{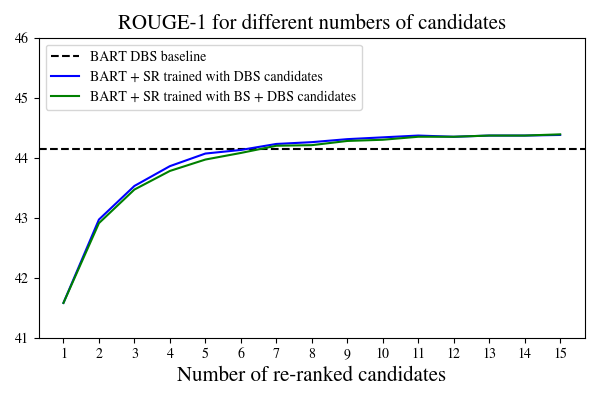}
\end{center}
\caption{\textbf{ROUGE-1 on XSum} for k sampled candidates at inference time, with $k \in \{ 1, \dots , 15 \}$. \textbf{SR} stands for SummaReranker, \textbf{BS} and \textbf{DBS} refer to beam search and diverse beam search, respectively. 
}
\label{fig:8}
\end{figure}

\begin{figure}
\begin{center}
\includegraphics[width=.4\textwidth]{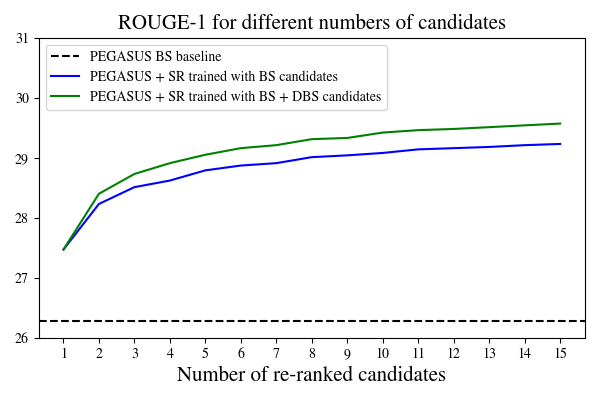}
\includegraphics[width=.4\textwidth]{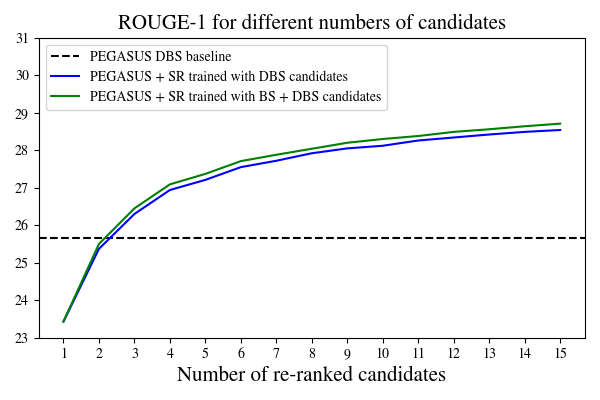}
\includegraphics[width=.4\textwidth]{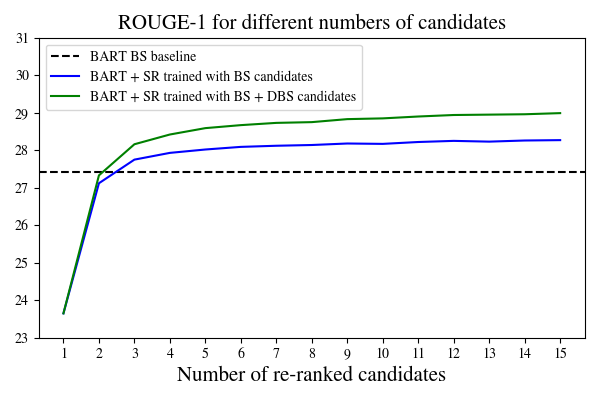}
\includegraphics[width=.4\textwidth]{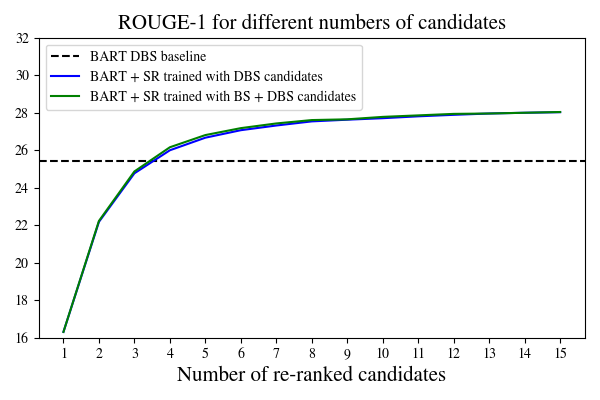}
\end{center}
\caption{\textbf{ROUGE-1 on Reddit TIFU} for k sampled candidates at inference time, with $k \in \{ 1, \dots , 15 \}$. \textbf{SR} stands for SummaReranker, \textbf{BS} and \textbf{DBS} refer to beam search and diverse beam search, respectively. 
}
\label{fig:9}
\end{figure}

In \Cref{fig:8}, we observe a failure mode of SummaReranker: on XSum and with PEGASUS when training the re-ranking with beam search candidates, performance decreases. However, the problem vanishes when SummaReranker is trained on a mixture of beam search and diverse beam search candidates. \\

\Cref{fig:9} top left (PEGASUS with beam search) represents a curious case: re-ranking a \emph{single} candidate is better than the top beam baseline. Since re-ranking a single candidate is equivalent to randomly sampling one candidate, this means that the top beam baseline is on average \emph{lower} than sampling a random candidate. We observed that such cases are rare and usually the top beam baseline is better than the random baseline. When the top beam baseline is lower, it is of utmost importance to keep all candidate and use a second-stage method to identify a better one.

%% file: Sections/appendix_k.tex
\begin{table}[h]
\centering
\resizebox{0.95\textwidth}{!}{
\begin{tabular}{lll}
\toprule
\multicolumn{3}{c}{CNN/DM} \\
\midrule
\multicolumn{2}{l}{Source} & \begin{tabular}[c]{@{}l@{}}Is this confirmation that Angel Di Maria is happy as a Manchester United player? The 27-year-old has endured a mixed start to his United \\ career on-and-off the pitch since joining the club last summer - which has included an attempted burglary at his family home in Cheshire back \\ in February. The midfielder has been linked with a move away from Old Trafford as a result, but speculation about his future could be squashed\\ following his latest tattoo. Angel Di Maria (left) has a new No 7 tattoo which stands out among others on his left arm . Di Maria wears the No 7 \\ shirt at Manchester United following his £60million from Real Madrid last summer. A new picture has been revealed on Twitter of Di Maria's \\ latest piece of body art - the number seven which stands out strongly among others on his left arm. United's club record £60million signing \\ of course adorns the No 7 shirt at the Red Devils - so could his latest tattoo suggest he's committed to Louis van Gaal's side for the long haul? \\ However, before United fans get too carried away it must be noted that the former Real Madrid star does also wear the No 7 jersey for Argentina\\ too. As well as adorning the No 7 shirt at United, 27-year-old (right) also wears that number for Argentina too.\end{tabular} \\
\midrule
Beam \#1 & Summary & \begin{tabular}[c]{@{}l@{}}Angel Di Maria has revealed his latest tattoo on Twitter. The 27-year-old has the No 7 shirt at Manchester United on his left arm. The Argentine has\\ endured a mixed start to his United career. He has been linked with a move away from Old Trafford as a result.\end{tabular} \\
& Reference scores & R-1: 38.6364,  R-2: 18.6047, R-L: 34.0909 // Rank: 15 \\
& Re-ranking       & SummaReranker score: 0.1577 // SummaReranker score rank: 15 \\
\hdashline
Beam \#2 & Summary & \begin{tabular}[c]{@{}l@{}}Angel Di Maria has a new tattoo of the No 7 shirt at Manchester United. The 27-year-old has endured a mixed start to his United career. The midfielder \\ has been linked with a move away from Old Trafford. Di Maria also wears the No 7shirt for Argentina too.\end{tabular} \\
& Reference scores & R-1: 59.0909,  R-2: 34.8837, R-L: 56.8182 // Rank: 5 \\
& Re-ranking       & SummaReranker score: 0.3905 // SummaReranker rank: 6 \\
\hdashline
Beam \#3 & Summary & \begin{tabular}[c]{@{}l@{}}Angel Di Maria has had a new No 7 tattoo on his left arm. The number stands out strongly among others on his arm. The 27-year-old joined Manchester \\ United for a club record £60million last summer. Di Maria also wears the No 7 shirt for Argentina.\end{tabular} \\
& Reference scores & R-1: 61.1765,  R-2: 33.7349, R-L: 58.8235 // Rank: 4 \\
& Re-ranking       & SummaReranker score: 0.4447 // SummaReranker rank: 5 \\
\hdashline
Beam \#4   & Summary          & \begin{tabular}[c]{@{}l@{}}Manchester United's record signing has a new No 7 tattoo on his left arm. Angel Di Maria wears the number seven shirt at Old Trafford. The 27-year-old \\ has endured a mixed start to his United career. He has been linked with a move away from Old Trafford as a result.\end{tabular} \\
& Reference scores & R-1: 37.7778,  R-2: 15.9091, R-L: 37.7778 // Rank: 14 \\
& Re-ranking       & SummaReranker score: 0.2528 // SummaReranker score: 10  \\
\hdashline
Beam \#5   & Summary  & \begin{tabular}[c]{@{}l@{}}Angel di Maria's latest tattoo shows him with the No 7 shirt at Manchester United. The 27-year-old has endured a mixed start to his United career. \\ The midfielder has been linked with a move away from Old Trafford. Di Maria joined United for a club record £60million from Real Madrid.\end{tabular} \\
& Reference scores & R-1: 53.3333,  R-2: 27.2727, R-L: 48.8889 // Rank: 10 \\
& Re-ranking       & SummaReranker score: 0.2377 // SummaReranker rank: 12 \\
\hdashline
Beam \#6   & Summary          & \begin{tabular}[c]{@{}l@{}}Argentina star Angel Di Maria has a new No 7 tattoo on his left arm. The number stands out strongly among others on his arm. Di Maria joined\\ Manchester United for a club record £60million last summe\$. The 27-year-old does also wear the No 7 shirt for Argentina too.\end{tabular} \\
& Reference scores & R-1: 61.3636,  R-2: 37.2093, R-L: 59.0909 // Rank: 2 \\
& Re-ranking       & SummaReranker score: 0.3058 // SummaReranker rank: 8 \\
\hdashline
Beam \#7   & Summary          & \begin{tabular}[c]{@{}l@{}}Manchester United's Angel Di Maria has had a new No 7 tattoo on his left arm. The 27-year-old's latest body art was revealed on Twitter. \\ Di Maria wears the No 7 shirt at Old Trafford following his £60million move from Real Madrid last summer.\end{tabular} \\
& Reference scores & R-1: 56.4706,  R-2: 31.3253, R-L: 47.0588 // Rank: 8 \\
& Re-ranking       & SummaReranker score: 0.8853 // SummaReranker rank: 2 \\
\hdashline
Beam \#8   & Summary          & \begin{tabular}[c]{@{}l@{}}The Manchester United star has revealed his latest tattoo on Twitter. Angel Di Maria has been linked with a move away from Old Trafford \\ in recent weeks. Di Maria wears the No 7 shirt at United following his £60million move from Real Madrid last summer.\end{tabular} \\
& Reference scores & R-1: 48.7805,  R-2: 25.0000, R-L: 43.9024 // Rank: 12 \\
& Re-ranking       & SummaReranker score: 0.2473 // SummaReranker rank: 11 \\
\hdashline
Beam \#9   & Summary          & \begin{tabular}[c]{@{}l@{}}\textbf{Manchester United's Angel Di Maria has had a new No 7 tattoo on his left arm. The 27-year-old's latest body art was revealed on Twitter.} \\ \textbf{Di Maria wears the No 7 shirt at Old Trafford following his £60million move from Real Madrid last summer. The Argentine also wears the number} \\ \textbf{for Argentina too.}\end{tabular} \\
& Reference scores & R-1: 61.7021,  R-2: 34.7826, R-L: 53.1915 // Rank: 6 \\
& Re-ranking       & \textbf{SummaReranker score: 0.9135} // \textbf{SummaReranker rank: 1} \\
\hdashline
Beam \#10  & Summary          & \begin{tabular}[c]{@{}l@{}}The Manchester United star has revealed his latest tattoo on Twitter. Angel Di Maria has been linked with a move away from Old Trafford in recent weeks. \\ Di Maria wears the No 7 shirt at United following his £60million move from Real Madrid last summer. The Argentine also wears the number for Argentina too.\end{tabular} \\
& Reference scores & R-1: 54.9451,  R-2: 29.2135, R-L: 50.5495 // Rank: 9 \\
& Re-ranking       & SummaReranker score: 0.1829 // SummaReranker rank: 14 \\
\hdashline
Beam \#11  & Summary          & \begin{tabular}[c]{@{}l@{}}Man Utd star Angel Di Maria has revealed his latest tattoo on Twitter. The 27-year-old has the No 7 shirt at Manchester United on his left arm. \\ Di Maria joined United for a club record £60million from Real Madrid last summer. The Argentine also wears the No 7 shirt for Argentina too.\end{tabular} \\
& Reference scores & R-1: 68.1319,  R-2: 40.4494, R-L: 61.5385 // Rank: 1 \\
& Re-ranking       & SummaReranker score: 0.3383 // SummaReranker rank: 7 \\
\hdashline
Beam \#12  & Summary          & \begin{tabular}[c]{@{}l@{}}Manchester United star Angel Di Maria has had a new No 7 tattoo. The number stands out strongly among others on his left arm. \\ Di Maria wears the No 7 shirt at Old Trafford following his £60million move. The Argentine also wears the number for his country too.\end{tabular} \\
& Reference scores & R-1: 54.1176,  R-2: 24.0964, R-L: 42.3529 // Rank: 11 \\
& Re-ranking       & SummaReranker score: 0.2172 // SummaReranker rank: 13 \\
\hdashline
Beam \#13  & Summary          & \begin{tabular}[c]{@{}l@{}}Manchester United midfielder Angel Di Maria has a new tattoo of the No 7 shirt at the club on his left arm. The 27-year-old has endured a mixed \\ start to his United career on-and-off the pitch since joining the club last summer. Di Maria has been linked with a move away from Old Trafford as a result.\end{tabular} \\
& Reference scores & R-1: 40.8163,  R-2: 20.8333, R-L: 36.7347 // Rank: 13 \\
& Re-ranking       & SummaReranker score: 0.2782 // SummaReranker rank: 9  \\
\hdashline
Beam \#14  & Summary          & \begin{tabular}[c]{@{}l@{}}Angel Di Maria has revealed his latest tattoo on Twitter. The 27-year-old has the number seven inked on his left arm.  Di Maria joined Manchester \\ United for a club record £60million last summer. The Argentine also wears the No 7 shirt for Argentina.\end{tabular} \\
& Reference scores & R-1: 58.5366,  R-2: 35.0000, R-L: 56.0976 // Rank: 7 \\
& Re-ranking       & SummaReranker score: 0.7447 // SummaReranker rank: 3 \\
\hdashline
Beam \#15  & Summary          & \begin{tabular}[c]{@{}l@{}}Angel di Maria has a new No 7 tattoo on his left arm. The number seven is among others on his left arm. The 27-year-old wears the No 7 shirt at \\ Manchester United. Di Maria joined United for a club record £60million from Real Madrid last summer.\end{tabular} \\
& Reference scores & R-1: 62.7907,  R-2: 33.3333, R-L: 58.1395 // Rank: 3 \\
& Re-ranking       & SummaReranker score: 0.4988 // SummaReranker rank: 4 \\
\midrule
\multicolumn{2}{l}{Reference} & \begin{tabular}[c]{@{}l@{}}Angel di Maria joined Manchester United from Real Madrid for £60million. Di Maria took the No 7 shirt upon his arrival at the English giants. \\ 27-year-old also wears the No 7 jersey for Argentina too.\end{tabular}   \\
\bottomrule
\end{tabular}
}
\caption{\textbf{Diverse beam search summary candidates} of a base PEGASUS and their ground truth and SummaReranker re-ranking scores on \textbf{CNN/DM}.}
\label{tab:20}
\end{table}

%%%%%%%%%%%%%%%%%%%%%%%%%%%%%%%%%%%%%%%%%%%%%%%%%%%%
\begin{table}[]
\centering
\resizebox{0.95\textwidth}{!}{
\begin{tabular}{lll}
\toprule
\multicolumn{3}{c}{XSum} \\
\midrule
\multicolumn{2}{l}{Source}    & \begin{tabular}[c]{@{}l@{}}Female officers will be able to wear a headscarf under their caps or berets, provided it is plain and is the same colour as the uniform. Headscarf bans on \\ university campuses and state institutions - except for the judiciary, military and police - have also been lifted in recent years. The garment has been \\ controversial in Turkey for years. Secularists regard it as a symbol of religious conservatism. Since the 1920s, Turkey has had a secular constitution \\ with no state religion. The opposition have accused President Recep Tayyip Erdogan and his Islamist-rooted Justice and Development Party (AKP) \\ of trying to reinterpret secularism. However, public debate has also evolved to accept the hijab as an expression of individual liberties, correspondents say. \\ No strong opposition has been voiced against this latest move. President Erdogan has long embraced Turks' right to express their religious beliefs openly, \\ but he says he is committed to secularism. In 2010, the country's universities abandoned an official ban on Muslim headscarves. Three years later, women \\ were allowed to wear headscarves in state institutions - with the exception of the judiciary, military and police. That year, four MPs wore headscarves \\ in parliament. Most people in Turkey are Sunni Muslims.\end{tabular} \\
\midrule
Beam \#1   & Summary          & The Turkish authorities have lifted a ban on female police officers wearing headscarves. \\
& Reference scores & R-1: 50.0000,  R-2: 27.2727, R-L: 41.6667 // Rank: 11 \\
& Re-ranking       & SummaReranker score: 0.6553 // SummaReranker rank: 12 \\
\hdashline
Beam \#2   & Summary          & Turkey has lifted a ban on female police officers wearing headscarves, the interior ministry says.\\
& Reference scores & R-1: 61.5385,  R-2: 41.6667, R-L: 61.5385 // Rank: 2 \\
& Re-ranking       & SummaReranker score: 0.8562 // SummaReranker rank: 2 \\
\hdashline
Beam \#3   & Summary          & The Turkish authorities have lifted a ban on female police officers wearing headscarves, state media report. \\
& Reference scores & R-1: 53.8462,  R-2: 25.0000, R-L: 53.8462 // Rank: 8 \\
& Re-ranking       & SummaReranker score: 0.5605 // SummaReranker rank: 1 \\
\hdashline
Beam \#4   & Summary          & Turkey has lifted its ban on female police officers wearing headscarves, the interior ministry says. \\
& Reference scores & R-1: 53.8462,  R-2: 25.0000, R-L: 53.8462 // Rank: 8 \\
& Re-ranking       & SummaReranker score: 0.7049 // SummaReranker rank: 9 \\
\hdashline
Beam \#5   & Summary          & The Turkish government has lifted a ban on female police officers wearing headscarves. \\
& Reference scores & R-1: 58.3333,  R-2: 36.3636, R-L: 50.0000 // Rank: 5 \\
& Re-ranking       & SummaReranker score: 0.7104 // SummaReranker rank: 8 \\
\hdashline
Beam \#6   & Summary          & The Turkish authorities have lifted a ban on police officers wearing headscarves. \\
& Reference scores & R-1: 52.1739,  R-2: 28.5714, R-L: 43.4783 // Rank: 10 \\
& Re-ranking       & SummaReranker score: 0.7503 // SummaReranker rank: 7 \\
\hdashline
Beam \#7   & Summary          & \textbf{Turkey has lifted a ban on female police officers wearing headscarves.} \\
& Reference scores & R-1: 63.6364,  R-2: 50.0000, R-L: 63.6364 // Rank: 1 \\
& Re-ranking       & \textbf{SummaReranker score: 0.9019 // SummaReranker rank: 1} \\
\hdashline
Beam \#8   & Summary          & Turkey's police force has lifted its ban on female officers wearing headscarves. \\
& Reference scores & R-1: 50.0000,  R-2: 18.1818, R-L: 50.0000 // Rank: 12 \\
& Re-ranking       & SummaReranker score: 0.6919 // SummaReranker rank: 10 \\
\hdashline
Beam \#9   & Summary          & Turkey's police force has lifted a ban on female officers wearing headscarves. \\
& Reference scores & R-1: 58.3333,  R-2: 36.3636, R-L: 58.3333 // Rank: 4 \\
& Re-ranking       & SummaReranker score: 0.8103 // SummaReranker rank: 5 \\
\hdashline
Beam \#10  & Summary          & Turkey's police force has lifted its ban on female officers wearing headscarves, officials say. \\
& Reference scores & R-1: 46.1538,  R-2: 16.6667, R-L: 46.1538 // Rank: 13 \\
& Re-ranking       & SummaReranker score: 0.5066 // SummaReranker rank: 15 \\
\hdashline
Beam \#11  & Summary          & The Turkish government has lifted a ban on female police officers wearing headscarves, state media report. \\
& Reference scores & R-1: 51.8519,  R-2: 32.0000, R-L: 44.4444 // Rank: 9 \\
& Re-ranking       & SummaReranker score: 0.6522 // SummaReranker rank: 13 \\
\hdashline
Beam \#12  & Summary          & Turkey's police force has lifted a ban on female officers wearing headscarves, state media report. \\
& Reference scores & R-1: 51.8519,  R-2: 32.0000, R-L: 51.8519 // Rank: 7 \\
& Re-ranking       & SummaReranker score: 0.7819 // SummaReranker rank: 6 \\
\hdashline
Beam \#13  & Summary          & Turkey has lifted its ban on female police officers wearing headscarves. \\
& Reference scores & R-1: 54.5455,  R-2: 30.0000, R-L: 54.5455 // Rank: 6 \\
& Re-ranking       & SummaReranker score: 0.8140 // SummaReranker rank: 4 \\
\hdashline
Beam \#14  & Summary          & Turkey has lifted a ban on female police officers wearing headscarves, the interior ministry has said. \\
& Reference scores & R-1: 59.2593,  R-2: 40.0000, R-L: 59.2593 // Rank: 3 \\
& Re-ranking       & SummaReranker score: 0.8298 // SummaReranker rank: 3 \\
\hdashline
Beam \#15  & Summary          & Turkey's police force has lifted its ban on female officers wearing headscarves, state media report.\\
& Reference scores & R-1: 44.4444,  R-2: 16.0000, R-L: 44.4444 // Rank: 15 \\
& Re-ranking       & SummaReranker score: 0.6728 // SummaReranker rank: 11 \\
\midrule
\multicolumn{2}{l}{Reference} & Turkey has lifted a ban on police women wearing the Islamic headscarf.   \\
\bottomrule
\end{tabular}
}
\caption{\textbf{Beam search summary candidates} of a base PEGASUS and their ground truth and SummaReranker re-ranking scores on \textbf{XSum}.}
\label{tab:21}
\end{table}

%%%%%%%%%%%%%%%%%%%%%%%%%%%%%%%%%%%%%%%%%%%%%%%%%%%%%%%%
\begin{table}[]
\centering
\resizebox{\textwidth}{!}{
\begin{tabular}{lll}
\toprule
\multicolumn{3}{c}{Reddit TIFU} \\
\midrule
\multicolumn{2}{l}{Source}    & \begin{tabular}[c]{@{}l@{}}here's my reconstruction of the fuck-up: during the visa application, i'm sifting through pages and pages of documentation with 15 tabs open on my browser \\ and i arrive at a page with the title english requirement. it says something like "here's a list of approved test providers and you have to score a minimum cefr\\  level of b1 to meet the english requirement." as someone who has taken many english exams such as toefl, ielts and pearson, i wonder what the hell a cefr level\\  is, how come i've never heard of this and start popping new pages. turns out you have to score that much from ielts or this much from pearson or that much \\ from other exams. cool. i'm thinking, currently i have 2 valid ielts exams that meet the criteria and a pearson's from which i've scored 90/90, sweet! i'll just submit\\  pearson's and done. so i pay 2000aud and get an appointment, submit my documents and come back home. "hey wifey, it was really easy, let's do the same for \\ you and get it done quickly." pay another 2000aud and my wife submits her application.  3 days after my submission, i get an e-mail saying a decision has been\\  made, yay? more like nay, refused because we don't accept pearson's. 2 days later, wife gets refuses as well because we refused your husband. 2000 down the \\ drain, luckily they are refunding the healthcare extras.  turns out i failed to go back to that webpage and click on the link to get access to the most ridiculous list \\ of approved english tests. there's not a mention of pearson, what's more, it doesn't have idp melbourne so my ielts exams are useless as well.  on the plus side, \\ i learn there are other ways to meet the requirement. if you have a diploma from an australian uni, just submit your diploma and you're good to go? why didn't \\ you give me a call or send me an email and say "hey, you made a mistake silly" and i'd be like here's my diploma or something, wtf is refusing? this will stick \\ to our international travel cv like a fucking bug.  plus, i had to ask my future employer for another certificate, which takes another two weeks and makes me\\  look like an idiot. then i'll have to submit another application, pay another 4000 and yada yada.  A bit of background: we are turkish citizens living in australia\\  as permanent residents obtained by using the same pearson's test. i'm working as a post-doctoral researcher and hopefully starting another post-doctoral position \\ in scotland, not giving up!\end{tabular} \\
\midrule
Beam \#1   & Summary          & i didn't go back to the page that says you have to score a cefr level of b1 to meet the english requirement. \\
& Reference scores & R-1: 4.4444,  R-2: 0.0000, R-L: 4.4444 // Rank: 12 \\
& Re-ranking       & SummaReranker score: 0.8117 // SummaReranker rank: 12 \\
\hdashline
Beam \#2   & Summary          & i'm an idiot. \\
& Reference scores & R-1: 0.0000,  R-2: 0.0000, R-L: 0.0000 // Rank: 15 \\
& Re-ranking       & SummaReranker score: 0.0612 // SummaReranker rank: 15 \\
\hdashline
Beam \#3   & Summary          & i didn't go back to the page that says you have to score a minimum cefr level of b1 to meet the english requirement. \\
& Reference scores & R-1: 4.3478,  R-2: 0.0000, R-L: 4.3478 // Rank: 13 \\
& Re-ranking       & SummaReranker score: 0.7396 // SummaReranker rank: 14 \\
\hdashline
Beam \#4   & Summary          & i didn't go back to the page that says you have to score a cefr level of b1 to meet the english requirement \\
& Reference scores & R-1: 4.4444,  R-2: 0.0000, R-L: 4.4444 // Rank: 11 \\
& Re-ranking       & SummaReranker score: 0.7940 // SummaReranker rank: 13 \\
\hdashline
Beam \#5   & Summary          & i didn't go back to the page where i found out i had to score a minimum cefr level of b1 to meet the english requirement. \\
& Reference scores & R-1: 4.1667,  R-2: 0.0000, R-L: 4.1667 // Rank: 14 \\
& Re-ranking       & SummaReranker score: 0.8320 // SummaReranker rank: 11 \\
\hdashline
Beam \#6   & Summary          & i didn't go back to the page that says you have to score a cefr level of b1 to meet the english requirement and now i'm stuck in australia. \\
& Reference scores & R-1: 7.6923,  R-2: 0.0000, R-L: 7.6923 // Rank: 7 \\
& Re-ranking       & SummaReranker score: 0.8954 // SummaReranker rank: 5 \\
\hdashline
Beam \#7   & Summary          & i didn't go back to the page that says you have to score a minimum cefr level of b1 to meet the english requirement and now i'm stuck in australia.\\
& Reference scores & R-1: 7.5472,  R-2: 0.0000, R-L: 7.5472 // Rank: 8 \\
& Re-ranking       & SummaReranker score: 0.8890 // SummaReranker rank: 8 \\
\hdashline
Beam \#8   & Summary          & i didn't go back to the page that says you have to score a minimum cefr level of b1 to meet the english requirement and now i'm stuck in a foreign country. \\
& Reference scores & R-1: 7.2727,  R-2: 0.0000, R-L: 7.2727 // Rank: 10 \\
& Re-ranking       & SummaReranker score: 0.8889 // SummaReranker rank: 9 \\
\hdashline
Beam \#9   & Summary          & i didn't go back to the page that says you have to score a cefr level of b1 to meet the english requirement on my visa application and now i'm stuck in australia. \\
& Reference scores & R-1: 7.2727,  R-2: 0.0000, R-L: 7.2727 // Rank: 10 \\
& Re-ranking       & SummaReranker score: 0.8941 // SummaReranker rank: 7 \\
\hdashline
Beam \#10  & Summary          & \textbf{\begin{tabular}[c]{@{}l@{}}i didn't go back to the page that says you have to score a minimum cefr level of b1 to meet the english requirement on my visa application and \\ now i'm stuck in australia.\end{tabular}} \\
& Reference scores & R-1: 21.4286,  R-2: 3.7037, R-L: 17.8571 // Rank: 1 \\
& Re-ranking       & \textbf{SummaReranker score: 0.9052 // SummaReranker rank: 1} \\
\hdashline
Beam \#11  & Summary          & \begin{tabular}[c]{@{}l@{}}i didn't go back to the page that says you have to score a minimum cefr level of b1 to meet the english requirement on my visa application and now i'm \\ stuck in australia.\end{tabular} \\
& Reference scores & R-1: 21.0526,  R-2: 3.6364, R-L: 17.5439 // Rank: 2 \\
& Re-ranking       & SummaReranker score: 0.9045 // SummaReranker rank: 2 \\
\hdashline
Beam \#12  & Summary          & \begin{tabular}[c]{@{}l@{}}i didn't go back to the page that says you have to score a minimum cefr level of b1 to meet the english requirement and now i'll have to submit another \\ application, pay 4000 and look like an idiot.\end{tabular} \\
& Reference scores & R-1: 12.9032,  R-2: 0.0000, R-L: 12.9032 // Rank: 6 \\
& Re-ranking       & SummaReranker score: 0.8861 // SummaReranker ran: 10 \\
\hdashline
Beam \#13  & Summary          & \begin{tabular}[c]{@{}l@{}}i didn't go back to the page that says you have to score a cefr level of b1 to meet the english requirement and now i'm going to have to submit another application \\ and pay 4000.\end{tabular} \\
& Reference scores & R-1: 13.5593,  R-2: 0.0000, R-L: 13.5593 // Rank: 3 \\
 & Re-ranking       & SummaReranker score: 0.8994 // SummaReranker rank: 3 \\
\hdashline
Beam \#14  & Summary          & \begin{tabular}[c]{@{}l@{}}i didn't go back to the page that says you have to score a minimum cefr level of b1 to meet the english requirement and now i'm going to have to submit another \\ application and pay 4000.\end{tabular} \\
& Reference scores & R-1: 13.3333,  R-2: 0.0000, R-L: 13.3333 // Rank: 4 \\
& Re-ranking       & SummaReranker score: 0.8947 // SummaReranker rank: 6 \\
\hdashline
Beam \#15  & Summary          & \begin{tabular}[c]{@{}l@{}}i didn't go back to the page that says you have to score a minimum cefr level of b1 to meet the english requirement and now i'm going to have to submit another \\ application and pay 4000 dollars.\end{tabular} \\
& Reference scores & R-1: 13.1148,  R-2: 0.0000, R-L: 13.1148 // Rank: 5 \\
& Re-ranking       & SummaReranker score: 0.8964 // SummaReranker rank: 4 \\
\midrule
\multicolumn{2}{l}{Reference} & made a silly mistake and got refused on 2x tier 2 uk visa applications for me and my partner costing 2000aud.      \\
\bottomrule
\end{tabular}
}
\caption{\textbf{Beam search summary candidates} of a base PEGASUS and their ground truth and SummaReranker re-ranking scores on \textbf{Reddit TIFU}.}
\label{tab:22}
\end{table}